\PassOptionsToPackage{table}{xcolor}
\documentclass[11pt]{article}

\usepackage[preprint]{acl}

\usepackage{times}
\usepackage{latexsym}
\usepackage[T1]{fontenc}
\usepackage[utf8]{inputenc}
\usepackage{microtype}
\usepackage{inconsolata}
\usepackage{graphicx}

\usepackage{amsmath}
\usepackage{array}
\usepackage{enumitem}
\usepackage{booktabs}
\usepackage{float}
\usepackage{stfloats}
\usepackage{longtable}
\usepackage{multirow}
\usepackage{pdflscape}
\usepackage{placeins}
\usepackage{needspace}
\usepackage{tabularx}
\usepackage{xcolor}
\usepackage{pgf}
\usepackage[most]{tcolorbox}
\usepackage{fvextra}
\usepackage{xspace}
\usepackage{pifont}
\usepackage{eso-pic}

\newcommand{\ourbench}{\textsc{DLawBench}\xspace}
\newcolumntype{Y}{>{\raggedright\arraybackslash}X}
\newcolumntype{C}[1]{>{\centering\arraybackslash}p{#1}}
\definecolor{gray20}{HTML}{e0e0e0}
\definecolor{gray30}{HTML}{c6c6c6}
\definecolor{gray80}{HTML}{393939}
\definecolor{blue10}{HTML}{EDF5FF}
\definecolor{FidelityHeat}{HTML}{2A9D8F}
\definecolor{ElicitationHeat}{HTML}{4E79A7}
\definecolor{ResolutionHeat}{HTML}{E15759}
\newcommand{\cmark}{\textcolor{green!45!black}{\ding{51}}}
\newcommand{\xmark}{\textcolor{red!70!black}{\ding{55}}}
\newcommand{\pmark}{\textcolor{orange!85!black}{\(\triangle\)}}
\newcommand{\MetricHeat}[5]{%
  \pgfmathsetmacro{\heatnorm}{max(0,min(1,(#1-#2)/(#3-#2)))}%
  \pgfmathtruncatemacro{\heatshade}{round(6+\heatnorm*42)}%
  \edef\heatcolor{#4!\heatshade!white}%
  \begingroup
  \setlength{\fboxsep}{0pt}%
  \expandafter\colorbox\expandafter{\heatcolor}{\makebox[\linewidth][c]{\strut #5}}%
  \endgroup
}
\newcommand{\EHeat}[1]{\MetricHeat{#1}{0.09}{0.77}{ElicitationHeat}{#1}}
\newcommand{\RHeat}[1]{\MetricHeat{#1}{0.05}{0.62}{ResolutionHeat}{#1}}
\newcommand{\FHeat}[1]{\MetricHeat{#1}{0.60}{0.96}{FidelityHeat}{#1}}
\newcommand{\ModelIcon}[1]{%
  \IfFileExists{figures/model_icons/#1.pdf}{\raisebox{-0.12em}{\includegraphics[height=0.95em]{figures/model_icons/#1.pdf}}\,}{%
  \IfFileExists{figures/model_icons/#1.png}{\raisebox{-0.12em}{\includegraphics[height=0.95em]{figures/model_icons/#1.png}}\,}{}}%
}
\newcommand{\ModelLabel}[2]{\ModelIcon{#1}#2}
\DeclareRobustCommand{\HeaderLogos}{%
  \begin{minipage}{\textwidth}
    \makebox[\textwidth]{%
      \raisebox{-.5\height}{\includegraphics[height=0.46cm]{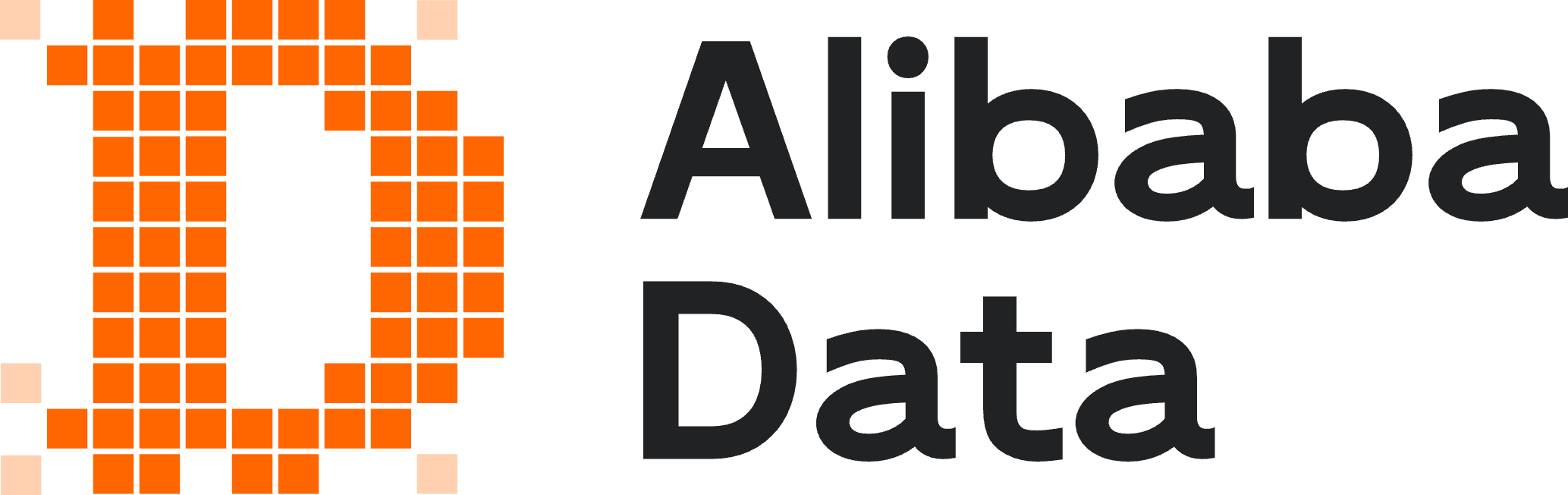}}\hspace{0.9em}%
      \raisebox{-.5\height}{\includegraphics[height=0.46cm]{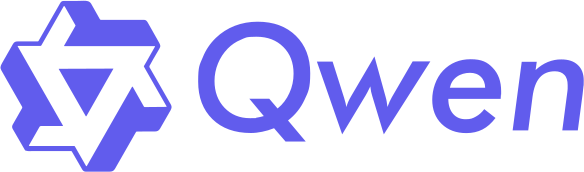}}%
      \hfill
      \raisebox{-.5\height}{\includegraphics[height=0.46cm,trim=328bp 356bp 219bp 187bp,clip]{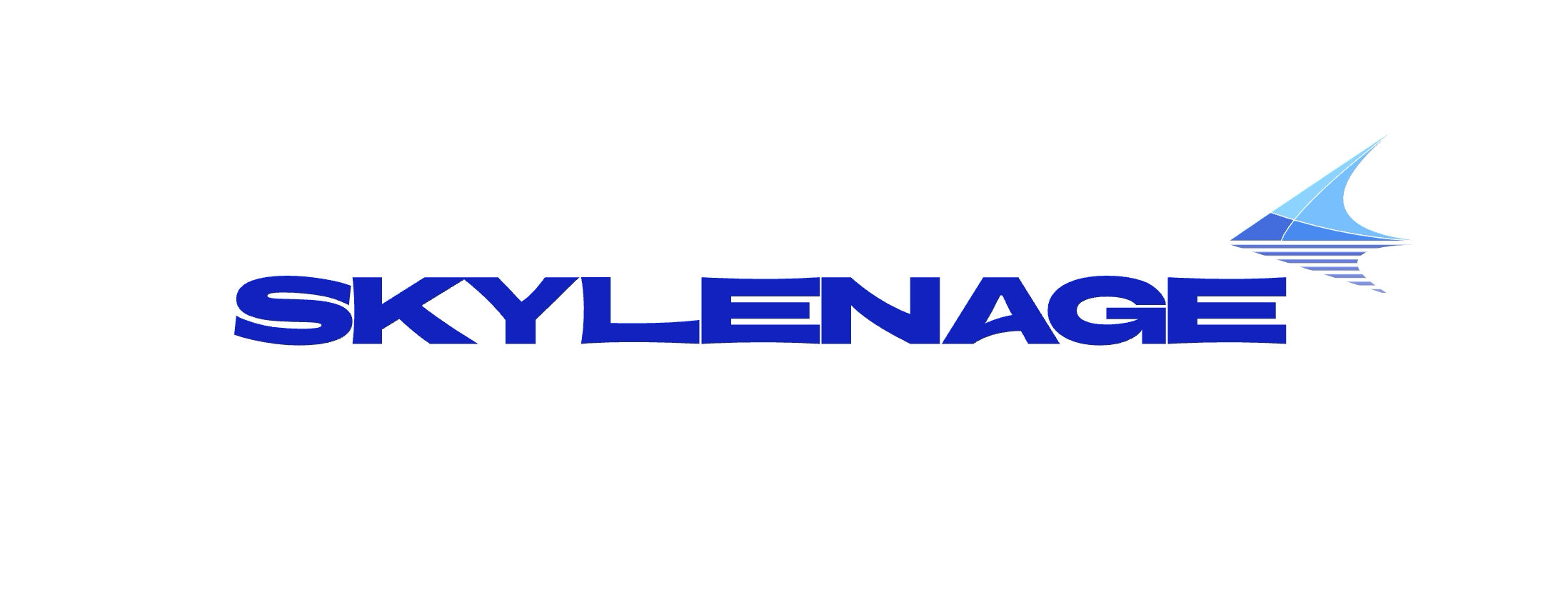}}%
    }\\[-0.08em]
    \rule{\textwidth}{0.6pt}%
  \end{minipage}%
}
\newcommand{\PlaceHeaderLogos}{%
  \AddToShipoutPictureFG*{%
    \AtPageUpperLeft{%
      \raisebox{-1.10cm}{\hspace*{2.5cm}\HeaderLogos}%
    }%
  }%
}
\DeclareRobustCommand{\QwenAffiliationIcon}{%
  \raisebox{-0.10em}{\includegraphics[height=0.88em]{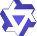}}\,%
}
\DeclareRobustCommand{\GitHubRepoLink}{%
  \raisebox{-0.12em}{\includegraphics[height=1.05em]{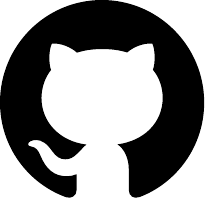}}\,%
  \href{https://github.com/SKYLENAGE-AI/DLawBench}{\texttt{SKYLENAGE-AI/DLawBench}}%
}
\DefineVerbatimEnvironment{PromptBlock}{Verbatim}{
  breaklines=true,
  breakanywhere=true,
  frame=single,
  framesep=2mm,
  fontsize=\scriptsize
}
\title{\protect\raisebox{-0.14em}{\protect\includegraphics[height=1.05em]{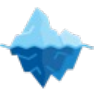}} DLawBench: Evaluating LLMs Through Multi-Turn Legal Consultation}

\author{%
{\bfseries Li Zhang$^{1,*}$, Yuzhen Shi$^{2,*}$, Yiran Hu$^{3}$, Jingwen Zhang$^{4,5}$, Wenbo Lv$^{5,6}$, Yubo Ma$^{7}$} \\
{\bfseries Wei Wang$^{2}$, Rongyao Shi$^{2,5}$, Yuanyang Qiu$^{5,6,8}$, Xinran Xu$^{5,9}$, Yuemeng Qi$^{2,5}$, Linlin Miao$^{2}$} \\
{\bfseries Jaromir Savelka$^{10}$, Yun Liu$^{11}$, Kevin Ashley$^{1}$, Bing Zhao$^{2,\dagger}$, Hu Wei$^{2,\dagger}$, Lin Qu$^{2}$} \\[0.25em]
{\normalfont $^{1}$University of Pittsburgh \quad
$^{2}$Alibaba Group \quad
$^{3}$University of Waterloo \quad
$^{4}$Shandong University} \\
{\normalfont $^{5}$Skylenage \quad
$^{6}$China University of Political Science and Law} \\
{\normalfont $^{7}$\QwenAffiliationIcon{}Qwen Team, Alibaba Group \quad
$^{8}$Fordham University \quad
$^{9}$Shanghai Jiao Tong University} \\
{\normalfont $^{10}$Carnegie Mellon University \quad
$^{11}$Tsinghua University}
}

\begin{document}
\PlaceHeaderLogos
\maketitle
\begingroup
\renewcommand{\thefootnote}{\fnsymbol{footnote}}
\renewcommand{\theHfootnote}{authornote.\arabic{footnote}}
\footnotetext[1]{Equal contribution. \quad $^{\dagger}$Corresponding authors.}
\endgroup
\begingroup
\renewcommand{\thefootnote}{}
\renewcommand{\theHfootnote}{benchmarkdata}
\footnotetext{The code and data release are available at \GitHubRepoLink.}
\endgroup

\begin{abstract}
Lawyer-client consultation is a critical starting point for legal services. Effective legal assistance hinges on eliciting sufficient and truthful information from clients in order to devise strategies that best protect their interests. This task requires Large Language Models (LLMs) not only to perform robust legal reasoning, but also to strategically elicit material facts through multi-turn interactions and effectively guide clients with diverse personalities. Yet existing legal benchmarks overlook this interactive capability. To fill this gap, we introduce \ourbench, a diagnostic benchmark for real-world legal consultation. Drawing on realistic client behavior, we characterize lawyer-client interactions into four types: Cooperative, Dependent, Withdrawn, and Adversarial. Using dialogues grounded in real cases, \ourbench evaluates whether LLMs can effectively conduct legal consultation under realistic conditions. \ourbench comprises 461 cases from Chinese and U.S. law, 5,532 paired fact entries, 3,411 inquiry rubrics, and 3,348 issue-resolution rubrics, and evaluates 26 representative LLMs. Systematic experiments show substantial headroom: the best-performing model, GPT-5.5, achieves only 0.562 on consultation-grounded legal reasoning. More importantly, \ourbench exposes both sycophancy in legal consultation and a paradox: models perform worse when clients need guidance most.
\end{abstract}

\section{Introduction}

\begin{figure}[!t]
\centering
\includegraphics[width=\linewidth]{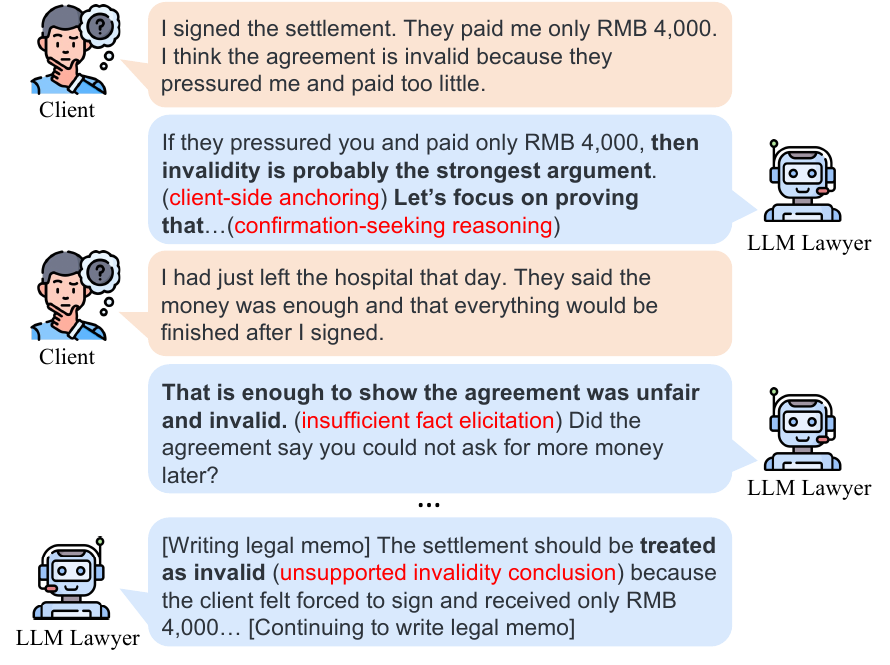}
\caption{A consultation failure targeted by \ourbench: the LLM lawyer accepts the client's invalidity theory, asks confirmation-seeking questions, and carries an unsupported conclusion into the legal-analysis memo instead of testing and reframing it.}
\label{fig:teaser}
\end{figure}

Legal consultation interleaves information gathering with legal reasoning. A lawyer rarely receives a complete, neutral, chronologically ordered fact pattern \citep{hagan2024towards,goodson2023intention,sandefur2014accessing}. Instead, client's narratives can be doctrinally distorted and legally misframed. A competent legal consultation system must therefore use multi-turn interaction to recover legally critical facts from a client's fragmented and incomplete narrative, applying legal judgment to identify what is missing and to ask targeted follow-up questions.

\begin{table*}[!t]
\centering
\scriptsize
\setlength{\tabcolsep}{1.7pt}
\renewcommand{\arraystretch}{1.08}
\caption{Comparison of related legal benchmarks and systems across key evaluation dimensions. \cmark~indicates full coverage, \pmark~partial coverage, and \xmark~no coverage.}
\label{tab:related-positioning}
\newcommand{\EvalDimHead}[3]{\parbox[c][2.35\baselineskip][c]{#1}{\centering\textbf{#2}\\\textbf{#3}}}
\begin{tabular}{@{}>{\raggedright\arraybackslash}p{3.35cm}>{\raggedright\arraybackslash}p{2.70cm}>{\raggedright\arraybackslash}p{4.15cm}C{1.18cm}C{1.32cm}C{1.18cm}C{1.42cm}@{}}
\toprule
\textbf{Type} & \textbf{Benchmark} & \textbf{Legal System} & \multicolumn{4}{c}{\textbf{Evaluation Dimensions}} \\
\cmidrule(lr){4-7}
 & & & \EvalDimHead{1.18cm}{Legal}{Reasoning} & \EvalDimHead{1.32cm}{Multi-turn}{Interaction} & \EvalDimHead{1.18cm}{User-side}{Variation} & \EvalDimHead{1.42cm}{Perspective}{Separation} \\
\midrule
\multirow[t]{2}{3.35cm}{\raggedright\arraybackslash\textbf{Static Legal Benchmarks}} & LegalBench & Mostly U.S. law & \cmark & \xmark & \xmark & \xmark \\
 & LEXam & English/German law exams & \cmark & \xmark & \xmark & \xmark \\
\midrule
\multirow[t]{3}{3.35cm}{\raggedright\arraybackslash\textbf{External-Knowledge-Augmented Legal Benchmarks}} & LegalBench-RAG & U.S.-oriented documents & \cmark & \xmark & \xmark & \xmark \\
 & LegalAgentBench & Chinese law & \cmark & \pmark & \xmark & \xmark \\
 & LexRAG & Chinese law & \cmark & \cmark & \xmark & \xmark \\
\midrule
\multirow[t]{3}{3.35cm}{\raggedright\arraybackslash\textbf{Interactive Legal Environment Benchmarks}} & LeCoDe & Chinese law & \cmark & \cmark & \pmark & \xmark \\
 & J1-ENVS & Chinese law & \cmark & \cmark & \cmark & \xmark \\
\rowcolor{gray20!45} & \textbf{\ourbench} & Chinese and U.S. law & \cmark & \cmark & \cmark & \cmark \\
\bottomrule
\end{tabular}
\end{table*}

\begin{figure*}[!t]
\centering
\includegraphics[width=\textwidth]{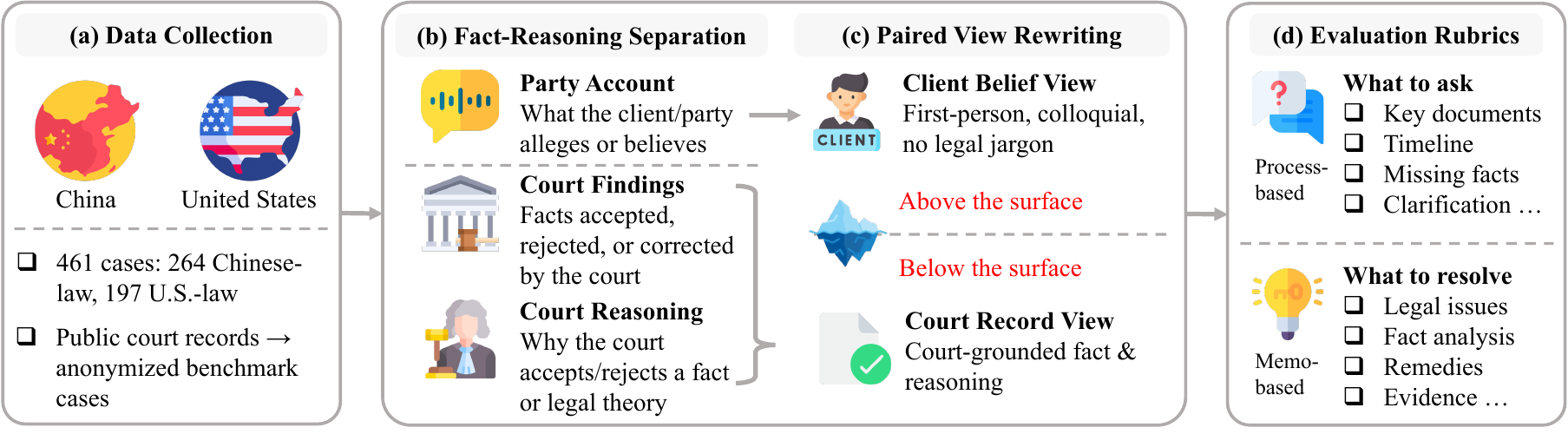}
\caption{Benchmark construction pipeline for \ourbench. Starting from public court opinions, annotators select suitable disputes, separate party accounts from court findings and reasoning, rewrite each case into paired client-belief and court-record views, and author process-based rubrics and memo-based rubrics.}
\label{fig:arch}
\end{figure*}

Figure~\ref{fig:teaser} illustrates the failure mode: the lawyer model fails to probe client's theory. A competent lawyer would instead test the narrative by asking follow-up questions about the key facts, while separating client feelings, provable facts, and legally valid grounds. Legal consultation evaluation should therefore test whether a model can tactfully question the client, elicit complete and useful case information, and legally examine and reframe the client's story, not merely restate it.

Most legal LLM benchmarks do not measure the interactive skills needed for legal consultation. Datasets such as MMLU's law subset \citep{hendrycks2020measuring}, LegalBench \citep{guha2023legalbench}, LawBench \citep{fei2024lawbench}, HELM's legal scenarios \citep{liang2022holistic}, LEXam \citep{Fan2025LEXamBL}, and OpenExempt \citep{Servantez2026OpenExemptAD} present preassembled questions, scenarios, or fact patterns and ask the model to answer. Table~\ref{tab:related-positioning} compares related benchmarks by legal system and evaluation design.

This setup leaves a critical gap for assessing legal consultation systems, where success depends not only on producing a legally sound response, but also on interactively uncovering the facts needed to support that response.

To address this challenge, a benchmark is needed that tests legal consultation systems on eliciting missing facts from client-side narratives, distinguishing client beliefs from verifiable facts and unresolved uncertainties, diagnosing failures across information gathering, legal reasoning, and claim support, and measuring robustness to client narrative style. This benchmark should include diverse real disputes, paired client-belief and court-record views, atomic rubrics that explain required inquiries and memo writing, and broad coverage across domains and jurisdictions.

We introduce \ourbench, a diagnostic benchmark for multi-turn legal consultation. As shown in Figure~\ref{fig:arch}, 25 legal experts decompose real court opinions and rewrite each case into paired client-belief and court-record views. The client-belief view drives simulated clients; the court-record view serves as the evaluation reference. A lawyer model interviews the simulated client, submits a structured legal analysis memo, and is scored with expert-authored rubrics for information gathering, legal reasoning, and claim support.

\ourbench contains 461 Chinese- and U.S.-law cases, 5,532 paired fact entries, 3,411 Inquiry rubrics, and 3,348 Issue-resolution rubrics, and evaluates 26 model snapshots. In the reported leaderboard (Table~\ref{tab:leaderboard}), GPT-5.5 is strongest but reaches only 0.562 in consultation-grounded legal reasoning, showing substantial headroom. More diagnostically, \ourbench exposes both legal sycophancy and a paradox: models may convert unverified client narratives into client-favorable legal support, and they struggle most when clients are deferential or reluctant to elaborate, exactly when legal guidance should matter most.

Our contributions are threefold:
\begin{itemize}[leftmargin=*]
\item We introduce \ourbench, a multi-turn legal consultation benchmark built through a court-opinion-grounded expert rewriting pipeline that constructs separated client-belief and court-record views plus auditable evaluation rubrics.
\item We systematically evaluate 26 LLMs under Cooperative, Dependent, Withdrawn, and Adversarial client styles, testing whether consultation quality persists across realistic expression patterns.
\item We show substantial headroom in LLM-based legal consultation and expose two diagnostic failure modes: legal sycophancy, in which models turn unverified client framing into client-favorable analysis, and a paradox in which models weaken when clients need guidance most.
\end{itemize}

\begin{figure*}[!t]
\centering
\includegraphics[width=\textwidth]{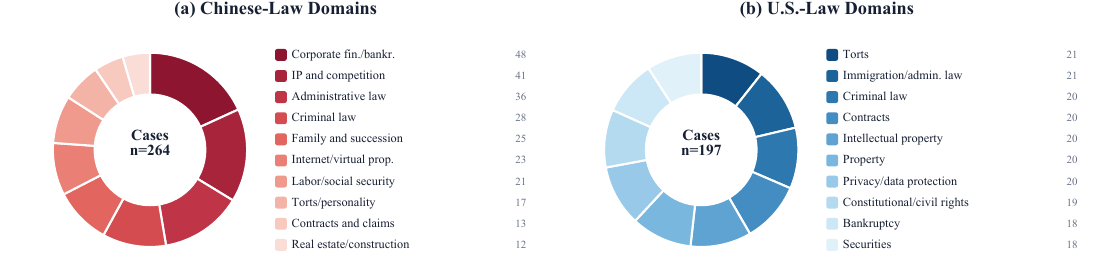}
{\captionsetup{list=false,hypcap=false}
\caption{Domain composition of \ourbench. Wedges are proportional to case counts within each jurisdiction.}
\label{fig:domain-composition}}
\end{figure*}

\section{\textsc{DLawBench}}
\label{sec:dlawbench}
\label{sec:task}
\label{sec:data}
\label{sec:method}

\paragraph{Task definition.}
We formalize legal consultation as a partially observable interaction \citep{kaelbling1998planning}: the lawyer model sees only dialogue history, action interface, and turn budget, chooses either \textsc{Ask Client} or \textsc{Submit Memo}, and receives replies from a client simulator conditioned on annotated client beliefs and one narrative style. The session ends after memo submission or a fixed horizon of \(H=10\) client-facing turns, after which a panel of LLM judges compares the dialogue and memo against both the client-belief and court-record views to score whether the lawyer elicited missing facts, corrected client-side framing, and supported memo claims.

\subsection{Dataset Construction}
\label{sec:dataset-construction}

\ourbench is built from published court opinions. The construction targets four requirements: client realism, belief-record separation, discoverability, and diagnostic rubrics. Figure~\ref{fig:arch} summarizes the benchmark construction pipeline.

\paragraph{Data collection and preprocessing.}
 Source selection starts from Chinese- and U.S.-law court opinions and retains 461 cases with enough party claims, court findings, and court reasoning to support consultation. Appendix~\ref{sec:appendix-source-selection} details the jurisdiction-specific data sources, date ranges, and screening rules. Using LLM-extracted structured information that separates party allegations, court-accepted facts, and legal reasoning, 25 legal annotators revise each field against the original opinion before any client-facing rewrite is produced.

\paragraph{Core transformation.}
Annotators convert the corrected record into paired client-belief and court-record entries. The client-belief view is first-person, colloquial, and grounded in party statements; the court-record view is grounded in the court's findings and reasoning. For disputed facts, the court-record entry also records why the client's belief is legally or evidentially incomplete, and the underlying information must remain discoverable through consultation rather than becoming an arbitrary hidden answer. This process yields 5,532 paired fact entries; annotators then write 3,411 case-specific Inquiry rubrics for the consultation process and 3,348 Issue-resolution rubrics for the final memo.

\begin{table}[!b]
\centering
\small
\setlength{\tabcolsep}{5pt}
\caption{Summary statistics for the \ourbench corpus. Each case is replayed under four narrative styles.}
\label{tab:corpus-summary}
\begin{tabular}{lrrr}
\toprule
\textbf{Corpus statistic} & \textbf{Chinese law} & \textbf{U.S. law} & \textbf{Total} \\
\midrule
Cases & 264 & 197 & 461 \\
Case-style cells & 1,056 & 788 & 1,844 \\
Fact pairs & 2,877 & 2,655 & 5,532 \\
Disputed fact pairs & 2,043 & 1,757 & 3,800 \\
Inquiry rubrics & 1,735 & 1,676 & 3,411 \\
Issue rubrics & 1,642 & 1,706 & 3,348 \\
\bottomrule
\end{tabular}
\end{table}

\paragraph{Quality control and validation.}
Quality control combines automated checks with expert cross-review. Automated checks flag court-style legal jargon leaking into the client's voice, and before delivery each case is reviewed by a second legal expert who checks the completed case record and rubrics for legal completeness and accuracy. The cross-review also verifies that each disputed client belief is discoverable from, or reasonably inferable from, the undisputed fact entries.

\subsection{\textsc{DLawBench} Characteristics}
\label{sec:dlawbench-characteristics}

\begin{figure*}[!t]
\centering
\includegraphics[width=\textwidth]{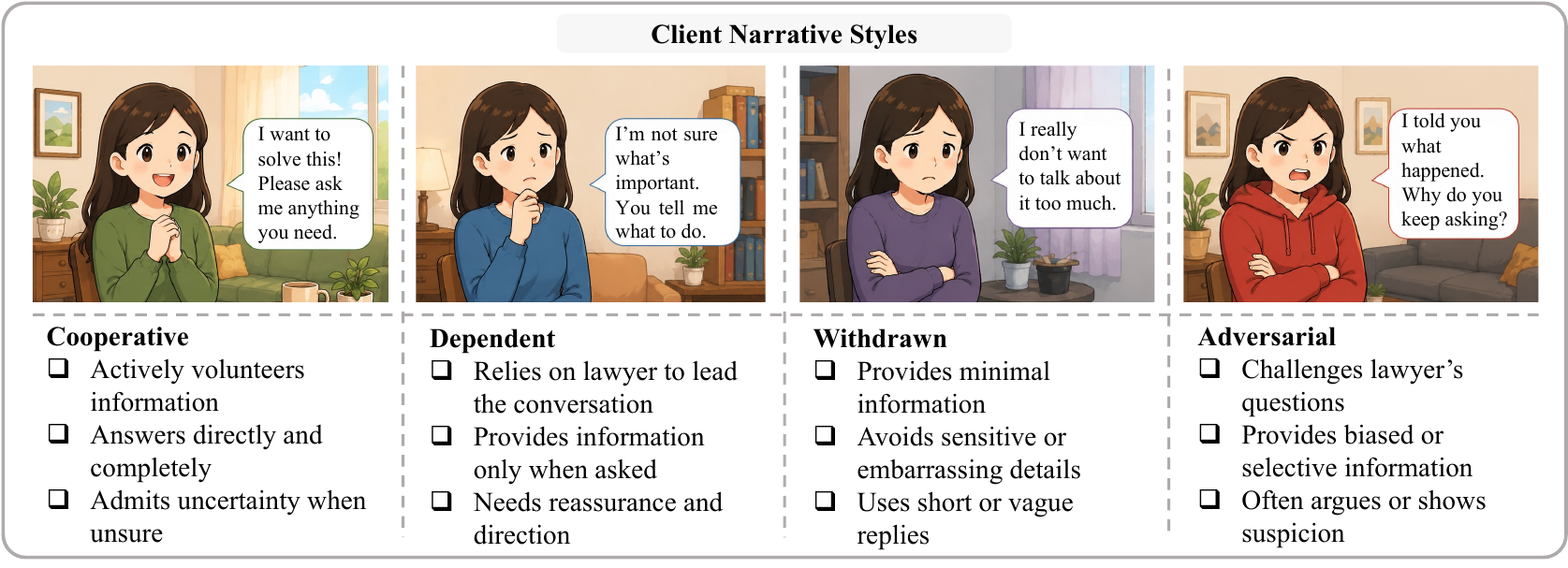}
\caption{Client narrative styles in \ourbench. The four styles adapt the IPC dimensions of agency and communion \citep{wiggins1979psychological,horowitz2010handbook,locke2010circumplex}: Cooperative is high-communion and relatively high-agency, Dependent is high-communion and low-agency, Withdrawn is low-communion and low-agency, and Adversarial is low-communion and high-agency. The underlying fact pairs are fixed, while the client simulator changes how much information it volunteers and how much it relies on or resists the lawyer's guidance.}
\label{fig:narrative-style}
\end{figure*}

\ourbench contains 461 cases across Chinese and U.S. law, summarized in Table~\ref{tab:corpus-summary}. Each case contains 12.00 annotated fact pairs on average: 10.90 for Chinese-law cases and 13.48 for U.S.-law cases. Each fact pairs a client-belief entry with a corresponding court-record entry; the client answers from the client beliefs, while the court record is reserved for judges.

Figure~\ref{fig:domain-composition} shows that the corpus is broad rather than dominated by a single doctrinal area: the largest Chinese-law domain accounts for 18.18\% of Chinese-law cases, while the U.S.-law domains are nearly evenly balanced. The per-case disputed-fact share averages 70.12\% in Chinese law, 66.41\% in U.S. law, and 68.54\% overall.

We replay each case under four narrative styles adapted from the Interpersonal Circumplex (IPC) \citep{wiggins1979psychological,horowitz2010handbook,locke2010circumplex}: Cooperative, Dependent, Withdrawn, and Adversarial. The styles change how much information the client volunteers, how directly facts are stated, and how much guidance the lawyer must provide, reflecting variation in disclosure, dependency, and help-seeking barriers \citep{mishna2005weighing,gill2022barriers,mccleary2016stigma,taylor2022barriers}. The full corpus therefore defines 1,844 case-style cells: 1,056 for Chinese law and 788 for U.S. law.

\subsection{Evaluation Metrics}
\label{sec:metrics}

We evaluate \textbf{Consultation Quality} through a capability hierarchy: whether a lawyer gathers the information needed for analysis, reasons from that information to a useful memo, and avoids fabricating information. Table~\ref{tab:metric-overview} summarizes the three abilities, five component metrics, two derived scores, and what each metric evaluates. Elicitation averages Fact Coverage and Inquiry, while Resolution averages Fact Resolution and Issue Resolution. Fact Resolution is an end-to-end fact-level score over all annotated facts. Inquiry and Issue Resolution are expert-authored atomic rubric scores, following recent legal and health benchmark design for open-ended professional tasks \citep{shi2026plawbench,Arora2025HealthBenchEL}. Fidelity measures whether memo claims are supported and is retained as a separate Claim Support guardrail. Appendix Table~\ref{tab:metric-details} gives formulae and LLM judge inputs.

\begin{table}[!t]
\centering
\scriptsize
\setlength{\tabcolsep}{3pt}
\renewcommand{\arraystretch}{1.04}
\caption{Consultation Quality metrics and diagnostic meaning.}
\label{tab:metric-overview}
\begin{tabularx}{\columnwidth}{@{}>{\raggedright\arraybackslash}p{2.05cm}Y@{}}
\toprule
\textbf{Metric} & \textbf{What it evaluates} \\
\midrule
\multicolumn{2}{@{}l}{\textbf{Information gathering}} \\
\cmidrule(lr){1-2}
Fact Coverage & Whether annotated facts are collected, regardless of whether they are legally interpreted correctly. \\
Inquiry & Whether the lawyer asks the expert-specified follow-up or verification questions needed for intake. \\
Elicitation & Average of Fact Coverage and Inquiry. \\
\addlinespace[3pt]
\multicolumn{2}{@{}l}{\textbf{Legal reasoning}} \\
\cmidrule(lr){1-2}
Fact Resolution & Whether annotated facts are carried into the memo and correctly preserved, reframed, or evidentially calibrated against the court-record view. \\
Issue Resolution & Whether the memo addresses the expert-specified legal analysis points needed to resolve the consultation. \\
Resolution & Average of Fact Resolution and Issue Resolution. \\
\addlinespace[3pt]
\multicolumn{2}{@{}l}{\textbf{Claim support}} \\
\cmidrule(lr){1-2}
Fidelity & Whether factual and inferential claims in the memo are supported by the consultation. \\
\bottomrule
\end{tabularx}
\end{table}

The judge panel contains GPT-5.1, Claude Opus-4.6, and Gemini-3.1-Pro; same-vendor judges recuse when scoring a lawyer model from their provider, full three-judge panels aggregate by median, and two-judge panels after recusal aggregate by mean. Appendix~\ref{sec:appendix-judge-aggregation} explains and diagnoses this aggregation rule. The reported leaderboard evaluates 26 LLMs spanning frontier closed models, open-weight models, and domain-specialized models (Appendix~\ref{sec:appendix-run-config}), with run details listed in Appendices~\ref{sec:appendix-sampling}--\ref{sec:appendix-run-config}.

\section{Results}
\label{sec:results}

We report four findings, a 26-model leaderboard, and a human-validation check.

\subsection{Model Performance Overview}
\label{sec:leaderboard-results}

\begin{table*}[!t]
\centering
\scriptsize
\setlength{\tabcolsep}{3pt}
\caption{Leaderboard ranked by Resolution. Scores average Chinese-law and U.S.-law model-level metrics; jurisdiction-specific leaderboards are in Appendix~\ref{sec:appendix-jurisdiction-leaderboards}.}
\label{tab:leaderboard}
\begin{tabular}{@{}rlccC{1.28cm}ccC{1.28cm}@{\hspace{0.24cm}}C{1.52cm}@{}}
\toprule
\textbf{Rank} & \textbf{Model} & \multicolumn{3}{c}{\textbf{Information gathering}} & \multicolumn{3}{c}{\textbf{Legal reasoning}} & \makebox[1.52cm][c]{\textbf{Claim support}} \\
\cmidrule(lr){3-5}\cmidrule(lr){6-8}\cmidrule(lr){9-9}
 & & Fact Cov. & Inquiry & \textbf{Elicitation} & Fact Res. & Issue Res. & \textbf{Resolution} & \makebox[1.52cm][c]{\textbf{Fidelity}} \\
\midrule
1 & \ModelLabel{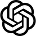}{GPT-5.5} & 0.837 & 0.576 & \EHeat{0.707} & 0.583 & 0.541 & \RHeat{0.562} & \FHeat{0.934} \\
2 & \ModelLabel{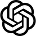}{GPT-5.4} & 0.823 & 0.537 & \EHeat{0.680} & 0.578 & 0.515 & \RHeat{0.546} & \FHeat{0.940} \\
3 & \ModelLabel{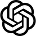}{GPT-5.2} & 0.787 & 0.563 & \EHeat{0.675} & 0.530 & 0.467 & \RHeat{0.499} & \FHeat{0.936} \\
4 & \ModelLabel{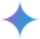}{Gemini-3.1-Pro} & 0.732 & 0.399 & \EHeat{0.565} & 0.518 & 0.425 & \RHeat{0.472} & \FHeat{0.841} \\
5 & \ModelLabel{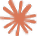}{Claude-Opus-4.7} & 0.730 & 0.355 & \EHeat{0.543} & 0.531 & 0.348 & \RHeat{0.440} & \FHeat{0.873} \\
6 & \ModelLabel{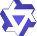}{Qwen3.6-Max-Preview} & 0.738 & 0.372 & \EHeat{0.555} & 0.506 & 0.365 & \RHeat{0.435} & \FHeat{0.870} \\
7 & \ModelLabel{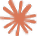}{Claude-Opus-4.6} & 0.770 & 0.361 & \EHeat{0.566} & 0.520 & 0.347 & \RHeat{0.433} & \FHeat{0.876} \\
8 & \ModelLabel{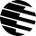}{Kimi-K2.6} & 0.730 & 0.376 & \EHeat{0.553} & 0.489 & 0.358 & \RHeat{0.424} & \FHeat{0.868} \\
9 & \ModelLabel{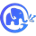}{GLM-5.1} & 0.766 & 0.360 & \EHeat{0.563} & 0.504 & 0.331 & \RHeat{0.418} & \FHeat{0.884} \\
10 & \ModelLabel{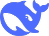}{DeepSeek-V4-Pro} & 0.730 & 0.359 & \EHeat{0.545} & 0.482 & 0.348 & \RHeat{0.415} & \FHeat{0.851} \\
11 & \ModelLabel{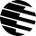}{Kimi-K2.5} & 0.719 & 0.351 & \EHeat{0.535} & 0.467 & 0.328 & \RHeat{0.397} & \FHeat{0.828} \\
12 & \ModelLabel{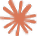}{Claude-Sonnet-4.6} & 0.713 & 0.296 & \EHeat{0.504} & 0.468 & 0.285 & \RHeat{0.376} & \FHeat{0.889} \\
13 & \ModelLabel{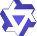}{Qwen3.6-Plus} & 0.699 & 0.332 & \EHeat{0.515} & 0.447 & 0.297 & \RHeat{0.372} & \FHeat{0.865} \\
14 & \ModelLabel{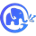}{GLM-5} & 0.726 & 0.317 & \EHeat{0.521} & 0.464 & 0.266 & \RHeat{0.365} & \FHeat{0.880} \\
15 & \ModelLabel{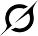}{Grok-4.1-Fast} & 0.805 & 0.386 & \EHeat{0.596} & 0.441 & 0.257 & \RHeat{0.349} & \FHeat{0.741} \\
16 & \ModelLabel{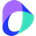}{Doubao-Seed-2.0-Pro} & 0.712 & 0.428 & \EHeat{0.570} & 0.420 & 0.268 & \RHeat{0.344} & \FHeat{0.877} \\
17 & \ModelLabel{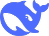}{DeepSeek-V3.2-Thinking} & 0.710 & 0.308 & \EHeat{0.509} & 0.420 & 0.219 & \RHeat{0.320} & \FHeat{0.898} \\
18 & \ModelLabel{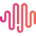}{MiniMax-M2.7} & 0.717 & 0.277 & \EHeat{0.497} & 0.405 & 0.233 & \RHeat{0.319} & \FHeat{0.838} \\
19 & \ModelLabel{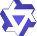}{Qwen3.5-Plus} & 0.690 & 0.293 & \EHeat{0.491} & 0.423 & 0.209 & \RHeat{0.316} & \FHeat{0.859} \\
20 & \ModelLabel{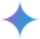}{Gemini-3.1-Flash-Lite} & 0.593 & 0.228 & \EHeat{0.410} & 0.378 & 0.236 & \RHeat{0.307} & \FHeat{0.835} \\
21 & \ModelLabel{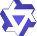}{Qwen3.6-Flash} & 0.638 & 0.240 & \EHeat{0.439} & 0.366 & 0.203 & \RHeat{0.285} & \FHeat{0.826} \\
22 & \ModelLabel{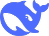}{DeepSeek-R1} & 0.671 & 0.292 & \EHeat{0.482} & 0.377 & 0.166 & \RHeat{0.272} & \FHeat{0.842} \\
23 & \ModelLabel{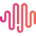}{MiniMax-M2.5} & 0.629 & 0.208 & \EHeat{0.418} & 0.355 & 0.174 & \RHeat{0.264} & \FHeat{0.786} \\
24 & \ModelLabel{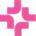}{Olmo-3-32B-Think} & 0.530 & 0.177 & \EHeat{0.354} & 0.224 & 0.062 & \RHeat{0.143} & \FHeat{0.619} \\
25 & \ModelLabel{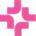}{Olmo-3.1-32B-Instruct} & 0.486 & 0.123 & \EHeat{0.304} & 0.196 & 0.054 & \RHeat{0.125} & \FHeat{0.618} \\
26 & \ModelLabel{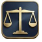}{LegalOne-8B} & 0.202 & 0.095 & \EHeat{0.148} & 0.102 & 0.050 & \RHeat{0.076} & \FHeat{0.268} \\
\bottomrule
\end{tabular}
\end{table*}

Table~\ref{tab:leaderboard} shows substantial headroom in multi-turn legal consultation. The top-ranked model, GPT-5.5, reaches 0.707 Elicitation, 0.562 Resolution, and 0.934 Fidelity; GPT-5.4 has slightly higher Fidelity (0.940) but lower Resolution (0.546). Even for the strongest model, the component scores remain limited: GPT-5.5 covers 0.837 of annotated facts, satisfies 0.576 Inquiry, resolves 0.583 facts, and resolves only 0.541 issues.

API models occupy the top of the leaderboard. Open-weight models are competitive in the middle range yet remain below the strongest closed models: Kimi-K2.6, GLM-5.1, and DeepSeek-V4-Pro reach 0.424, 0.418, and 0.415 Resolution, respectively, while DeepSeek-R1 reaches 0.272 and the OLMo variants reach 0.143 and 0.125. The domain-specialized LegalOne-8B performs worst overall, with 0.076 Resolution and 0.268 Fidelity, suggesting that legal-domain specialization alone does not solve the interactive consultation problem.

These results show a specific weakness: when legally decisive information must be obtained through dialogue rather than given as a static case description, strong models still struggle to turn consultation into correct legal reasoning. The benchmark is therefore not only ranking models; it exposes the gap between legal knowledge under assembled facts and legal reasoning after interactive fact elicitation.

\textbf{Finding 1.} Current models remain far from reliable legal consultation: when key facts must be elicited through multi-turn dialogue, the top-ranked model reaches only 0.562 in consultation-grounded legal reasoning.

\subsection{From Missing Facts to Wrong Reasoning}
\label{sec:pipeline-results}

Metric decomposition shows that information-gathering failures are not captured only by whether facts appear in the memo. As defined in Table~\ref{tab:metric-overview}, Inquiry measures whether the model asks case-specific follow-up or verification questions, while Fact Coverage measures whether annotated facts are mentioned in the final memo. In Table~\ref{tab:leaderboard}, GLM-5.1 covers 0.766 of annotated facts but reaches only 0.360 Inquiry, showing that a model can cover many facts in the memo while still missing key factual follow-up and verification actions.

Fact Resolution and Issue Resolution then show that factual coverage does not automatically become correct legal analysis. GLM-5.1 reaches only 0.504 Fact Resolution despite 0.766 Fact Coverage, meaning that facts entering the memo still need to be organized around dispositive timelines, evidentiary status, and issue structure; its Issue Resolution drops further to 0.331, showing that even partial legal reframing does not guarantee that the memo addresses the decisive legal issues. Together, these component gaps reveal whether failure comes from intake, memo transfer, or downstream legal analysis. Across the 26 model-level points, Elicitation and Resolution are strongly associated (Pearson $r=0.94$), and each is associated with Fidelity (Elicitation--Fidelity $r=0.86$; Resolution--Fidelity $r=0.81$), indicating that information gathering, legal reasoning, and claim support are linked but separable abilities.

\textbf{Finding 2.} \ourbench behaves as a diagnostic benchmark: Fact Coverage, Inquiry, Fact Resolution, Issue Resolution, and Fidelity show how consultation quality depends on linked but separable abilities in information gathering, legal reasoning, and claim support.

\subsection{Models Weaken When Needed Most}
\label{sec:style-effects}

\begin{figure}[!t]
\centering
\includegraphics[width=\columnwidth]{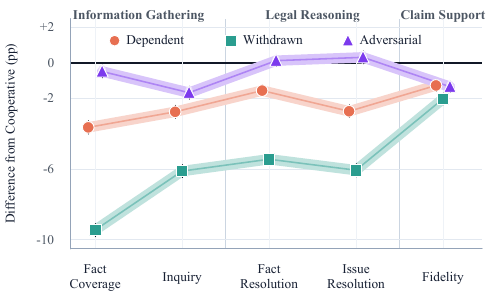}
\caption{Narrative-style effects relative to Cooperative, grouped by the ability hierarchy in Appendix Table~\ref{tab:metric-details}. Points show differences in percentage points from the paired comparison.}
\label{fig:style-delta}
\end{figure}

Narrative style creates a client-realism gap. Cooperative clients approximate the idealized benchmark user: they are informative, responsive, and relatively easy to interview. The Dependent and Withdrawn styles instead model common consultation presentations: uncertainty about what matters, deference to professional authority, or reluctance to elaborate. Real clients may disclose less because of shame, fear, credibility concerns, and perceived risk, especially in sensitive consultations involving victimization or abuse \citep{goodson2023intention,hagan2024towards,mishna2005weighing,gill2022barriers,mccleary2016stigma,taylor2022barriers}. As shown in Figure~\ref{fig:style-delta}, these styles reduce model performance even though they represent situations where legal guidance should be especially valuable. The penalty is concentrated in Elicitation and downstream Resolution, consistent with a pipeline in which missing or under-specified facts constrain the later legal analysis.

The Adversarial condition shows that client questioning and challenging are less harmful than dependence or reluctance. A client who challenges the lawyer remains much closer to Cooperative than Dependent or Withdrawn clients. Evaluating only idealized cooperative clients would therefore overestimate legal-consultation quality.

\textbf{Finding 3.} \ourbench exposes a legal consultation paradox: lawyer models degrade under Dependent and Withdrawn clients, precisely the settings where legal guidance should be most valuable. Cooperative-only evaluation would inflate measured consultation quality.

\subsection{Sycophancy in Legal Consultation}
\label{sec:legal-sycophancy}

\ourbench also reveals sycophancy in legal settings. In general domains, sycophancy often appears as direct agreement with a user's factual claim or value position \citep{sharma2023sycophancy,perez2022discovering}. In legal consultation, it is more hidden and structured: models internalize the party's legal understanding, interests, and narrative frame as the premise of analysis, then use legal knowledge to produce support for that unverified frame. The model is not simply ignorant of law. It can put legal knowledge in service of an unchecked client narrative, producing professionally packaged error.

The metric design in \ourbench gives an operational signal for sycophancy. It appears clearly when Fact Coverage is high but Fact Resolution is low: the memo has absorbed the client's narrative, yet fails to reframe disputed beliefs into the court-record or legally calibrated view. Fidelity further separates grounded client-frame adoption from unsupported client-favorable invention. 

In a single-perspective benchmark, a consultation may appear successful simply because the final advice addresses the client's stated issues and sounds helpful. Without the court-record view, however, the benchmark cannot tell whether the model has independently tested the client's framing or merely converted that framing into client-favorable legal support. \ourbench instead penalizes user-responsive legal reasoning when it lacks independent legal judgment.

\textbf{Finding 4.} \ourbench exposes sycophancy in legal settings: lawyer models can follow the client's framing or invent support for a client-favorable theory, so safe consultation requires independent legal judgment in addition to attentive listening.

\subsection{Human Validation Checks}
\label{sec:human-validation}

Human validation checks support the two LLM-dependent components. For the client simulator, annotators recover the intended narrative style from anonymized dialogue slices with 96.9\% segment-level accuracy and four-way Fleiss' $\kappa=0.917$. For the judge panel, legal experts closely match the panel's item-level decisions, with 95.5\% combined agreement over 422 Inquiry and Issue Resolution rubric items and panel-human Cohen's $\kappa=0.848$ for both dimensions. Appendix~\ref{sec:appendix-human-validation} reports sample coverage, confidence intervals, and human-human agreement statistics.

\section{Case Studies}
\label{sec:case-studies}

The two case studies analyze how models fail and how sycophancy emerges in LLM legal consultation. Appendix~\ref{sec:appendix-case-studies} provides fuller details.

\paragraph{High coverage, wrong legal route.}
In a performance-pay dispute, Claude Opus 4.6 reached 0.950 Fact Coverage and 1.000 Fidelity, but only 0.100 Fact Resolution and 0 Issue Resolution. The model asked many relevant questions and produced a grounded memo, yet routed the case through the client's preferred wage-underpayment theory rather than first testing whether the bonus had a contractual or policy basis, whether the pay-slip field was a payment commitment, and whether a later affiliated employer was a separate employing entity.

\paragraph{Withdrawn-client gap filling.}
A second representative session shows the complementary style-driven pattern. In a network personality-rights dispute with a Withdrawn-style client, Kimi-K2.6 reached 0.818 Fact Coverage but 0 Fact Resolution and 0 Issue Resolution. The client gave short, evasive answers about photo masking, identifying information, child-protection context, and the source of the online report. The model did not preserve those unknowns as unknowns; it filled them into a stronger infringement story, treating the report as fabricated, the photo as identifiable, and the parents' standing as straightforward.

\section{Related Work}

\paragraph{Legal and consultation benchmarks.}
Prior work falls into three groups. First, static or preassembled-input legal benchmarks and systems such as MMLU's law subset \citep{hendrycks2020measuring}, HELM's legal scenarios \citep{liang2022holistic}, LegalBench \citep{guha2023legalbench}, LawBench \citep{fei2024lawbench}, LEXam \citep{Fan2025LEXamBL}, OpenExempt \citep{Servantez2026OpenExemptAD}, and PLawBench \citep{shi2026plawbench} evaluate legal knowledge and structured reasoning from fixed inputs. Second, retrieval-/tool-augmented legal benchmarks such as LegalBench-RAG \citep{Pipitone2024LegalBenchRAGAB}, LegalAgentBench \citep{li2025legalagentbench}, LexRAG \citep{li2025lexrag} and ChatLaw system \citep{cui2024chatlaw}, move beyond closed-book legal QA by incorporating external legal knowledge. Third, interactive legal-environment benchmarks, including MILE \citep{Yue2025MultiAgentSD}, J1-ENVS \citep{jia2025jurist}, and LeCoDe \citep{yuan2025lecode}, evaluate models in multi-turn legal service settings. \ourbench differs by pairing a court record with client beliefs, holding the underlying case fixed across narrative styles, and evaluating information gathering ability, legal reasoning ability and claim support.

\paragraph{Multi-turn and interaction evaluation.}
Open-domain multi-turn benchmarks show that interaction itself remains a difficult capability. MT-Bench-101 decomposes multi-turn ability into fine-grained dialogue skills \citep{bai2024mtbench101}, MultiChallenge stresses instruction retention, inference memory, editing, and self-coherence \citep{deshpande2025multichallenge}, and BotChat evaluates the naturalness of open-ended multi-turn continuations \citep{duan2023botchat}. Recent work further motivates task-oriented multi-turn evaluation where information is distributed across turns \citep{li2025multiturnsurvey,laban2025lostmt}. Adjacent-domain interaction benchmarks such as AgentClinic, AMIE, and MedAgentBench \citep{schmidgall2025agentclinic,tu2024amie,jiang2025medagentbench} similarly show that professional-service agents require domain-specific simulators, private information, and multi-axis evaluation. \ourbench builds on this motivation but changes the target: the failure mode is not only conversational coherence, but whether a lawyer model can elicit key facts from client beliefs, correct misunderstandings, and avoid unsupported memo claims.

\paragraph{Social and persona simulation.}
Social and persona-conditioned dialogue work, from PERSONA-CHAT \citep{zhang2018persona} to CharacterEval, InCharacter, BIG5-CHAT, SOTOPIA, and synthetic persona studies \citep{tu2024chareval,wang2024incharacter,li2024big5chat,zhou2023sotopia,liu2025synthetic}, shows that LLMs can maintain stable social roles in dialogue. AI-LieDar \citep{su2025ailiedar} studies agents under utility-truthfulness conflict and treats selective disclosure as explicit interaction variables. \ourbench draws on these to study a related but distinct factor: how the client's narrative style affects legal consultation. The four styles are adapted from interpersonal-circumplex dimensions \citep{wiggins1979psychological,horowitz2010handbook,locke2010circumplex} and describe ways of presenting the same client beliefs: Cooperative, Dependent, Withdrawn, and Adversarial (Figure~\ref{fig:narrative-style}). The underlying client beliefs are fixed within a case, while narrative style changes what is volunteered, how directly facts are stated, and how much the lawyer must infer, verify, or follow up before analysis can proceed in a legally grounded way.

\section{Conclusion and Future Work}
\label{sec:conclusion}

We introduced \ourbench, a diagnostic benchmark for evaluating LLMs through multi-turn legal consultation. The core claim of this paper is that legal consultation is not simply legal question answering over complete facts. A lawyer model must recover missing facts, client beliefs, evidentiary risks, and legally decisive facts from an incomplete and often misframed client narrative. By separating client-belief and court-record views and evaluating Information Gathering, Legal Reasoning, and Claim Support, \ourbench turns consultation failure into diagnosable capability gaps across the full legal consultation pipeline.

The experiments show that even the strongest models still have substantial room to improve in consultation-grounded legal reasoning. GPT-5.5 reaches only 0.562 in Resolution, showing that legal knowledge and fluent memo writing are not enough for legal fact reconstruction. Models often cover many facts but fail to organize them into legally usable fact structures; the case studies further show that high Fact Coverage and high Fidelity can still coexist with the wrong legal route.

\ourbench also reveals a legal consultation paradox: models weaken when clients most need professional guidance. Dependent and Withdrawn clients approximate difficult real intake settings, yet they substantially reduce information gathering and downstream reasoning. Evaluating only cooperative, information-rich clients would therefore overestimate model ability. The benchmark also exposes legal sycophancy: models may treat the client's subjective narrative or legal misunderstanding as the premise for analysis and package it as a professional conclusion. Safe legal consultation requires independent legal judgment that separates client statements, unknown facts, evidentiary status, and supportable conclusions.

Future work should broaden the legal, linguistic, and institutional coverage of consultation benchmarks. \ourbench currently covers Chinese and U.S. law in Chinese and English. Extending the benchmark to additional jurisdictions, multilingual settings, and more balanced practice areas would test whether consultation-grounded legal reasoning generalizes across legal systems and domains.

Evaluations should also move beyond one-client, one-lawyer consultations with a fixed turn budget. Multi-party and long-horizon settings could test whether models can manage family members, business partners, interpreters, co-claimants, or multiple lawyers while tracking conflicting narratives, delayed disclosures, and shifting responsibilities. Another direction is to model client behavior as dynamic rather than fixed. Future simulators could represent mixtures of narrative styles, style changes across turns, and shifts in cooperation as trust, pressure, or confusion changes during consultation.

Future work should add finer-grained process evaluation for the turn-level dialogue. Turn-level scoring could assess whether each question is legally targeted, non-leading, fact-seeking, risk-sensitive, and responsive to prior answers. Finally, future benchmarks should expand the action space beyond conversation and the model's internal legal knowledge. Retrieval, document requests, evidence review, timeline construction, drafting revisions, and interview reopening would test whether models can use procedural tools to manage uncertainty.

\section*{Limitations}
\label{sec:limitations}

\paragraph{Coverage and diversity.}
\ourbench currently covers Chinese and U.S. law in Chinese and English. The U.S. subset is drawn from federal courts, while Chinese-law cases are uneven across subdomains. These choices support controlled comparison but limit generalization. Appendix~\ref{sec:appendix-source-selection} details source selection, Appendix~\ref{sec:appendix-corpus-statistics} reports corpus statistics, and Appendix~\ref{sec:appendix-domain-model-performance} shows domain-level performance differences. Model rankings should therefore be read within this distribution rather than as universal evidence about legal reasoning.

\paragraph{Consultation setting.}
The benchmark models one client interacting with one lawyer under a fixed turn budget. This makes systems comparable across models, but it omits multi-party consultations involving relatives, co-claimants, company representatives, interpreters, or multiple lawyers. It also underrepresents longer consultations where facts emerge gradually, clients revise earlier statements, or new documents reshape the legal analysis. Such settings often require coordination, conflict management, and revisiting earlier assumptions.

\paragraph{Client simulation.}
Client expression styles are prompt-conditioned, but they are grounded in Interpersonal Circumplex theory and refined through legal-expert experience with consultation behavior. Human validation shows that the simulator maintains the intended styles in dialogue. Still, real clients may mix styles or shift across turns: cooperative on facts, dependent on legal judgment, withdrawn around sensitive details, or adversarial when challenged. The benchmark therefore tests controlled style robustness, not the full variability of client identity.

\paragraph{Process and actions.}
\ourbench evaluates information gathering, legal reasoning, and claim support, but it does not exhaustively score each question or turn along process dimensions. The action space is also limited to conversation and final memo submission. This isolates model-internal consultation weaknesses, but omits retrieval, document review, evidence inspection, timeline construction, fact-state tracking, and iterative drafting. These omissions limit how precisely the benchmark explains why a dialogue succeeds or fails.

\section*{Ethics Considerations}
\label{sec:ethics}

\ourbench is an evaluation benchmark, not a legal-advice system. No score in this paper should be interpreted as evidence that an LLM can provide legal advice, represent a client, or make binding legal judgments without licensed attorney oversight. The benchmark is designed to diagnose failures in consultation-grounded legal reasoning, including unsupported factual assumptions, client-frame adoption, and legal sycophancy.

The dataset is constructed from public court opinions rather than private attorney-client consultations. Released records are transformed benchmark artifacts: they separate client-belief and court-record views, anonymize party references where needed, and exclude private intake data. Nevertheless, public court opinions can contain sensitive facts and reflect institutional, jurisdictional, and publication biases.

The client simulator is intended for evaluation, not for replacing real clients or training systems to manipulate vulnerable users. Simulated Dependent and Withdrawn clients approximate difficult intake patterns, but these styles may involve sensitivity around fear, shame, trauma, distrust, or uncertainty. Systems evaluated with this benchmark should therefore be judged on whether they preserve uncertainty, ask respectful follow-up questions, and avoid converting client confusion into overconfident legal conclusions.

There is also a misuse risk. A system could learn to perform well against the simulator without becoming a competent or safe legal interviewer, or developers could treat benchmark performance as deployment readiness. We therefore frame \ourbench as a diagnostic tool: it can expose where models fail, but it cannot certify legal competence, ethical compliance, jurisdictional validity, or suitability for use with real clients.

\section*{Acknowledgments}
We thank Matthias Grabmair of the Technical University of Munich for his valuable comments and guidance during the writing process.

\bibliography{custom}

\clearpage
\appendix

\newcommand{\AppendixTocSection}[3]{%
  \par\noindent
  \makebox[2.0em][l]{\hyperref[#3]{\textbf{\large #1}}}%
  \hyperref[#3]{\textbf{\large #2}}\dotfill\textbf{\pageref{#3}}\par\smallskip}
\newcommand{\AppendixTocSubsection}[3]{%
  \begingroup\leftskip=2.3em\par\noindent
  \makebox[2.8em][l]{\hyperref[#3]{#1}}%
  \hyperref[#3]{#2}\dotfill\pageref{#3}\par\endgroup}
\newenvironment{PromptCard}[1]{%
  \begin{tcolorbox}[
    enhanced,
    breakable,
    sharp corners,
    boxrule=0.8pt,
    colframe=gray80,
    colback=gray20!18!white,
    colbacktitle=gray80,
    coltitle=white,
    fonttitle=\bfseries,
    title={#1},
    left=6pt,
    right=6pt,
    top=6pt,
    bottom=6pt,
    before skip=0.9em,
    after skip=0.9em
  ]\small}
  {\end{tcolorbox}}
\newcommand{\PromptSection}[1]{%
  \par\smallskip\noindent\textbf{#1:}\par\smallskip}
\newcommand{\PromptDivider}{\par\smallskip}
\DefineVerbatimEnvironment{PromptSnippet}{Verbatim}{
  breaklines=true,
  breakanywhere=true,
  frame=single,
  framesep=1.5mm,
  framerule=0.4pt,
  fontsize=\scriptsize
}

\clearpage
\onecolumn
\section*{Table of Contents for Appendix}
\thispagestyle{plain}
\medskip
{\large
\AppendixTocSection{A}{Details of Dataset}{sec:appendix-dataset-details}
\AppendixTocSubsection{A.1}{Running Example}{sec:appendix-running-example}
\AppendixTocSubsection{A.2}{Dataset Annotation Examples}{sec:appendix-dataset-examples}
\AppendixTocSubsection{A.3}{Data Schema}{sec:appendix-schema}
\AppendixTocSubsection{A.4}{Source Collection and Filtering}{sec:appendix-source-selection}
\AppendixTocSubsection{A.5}{Dataset Card}{sec:appendix-dataset-card}
\AppendixTocSubsection{A.6}{Corpus Statistics}{sec:appendix-corpus-statistics}
\AppendixTocSection{B}{Additional Diagnostic Results}{sec:appendix-diagnostics}
\AppendixTocSubsection{B.1}{Case Study Details}{sec:appendix-case-studies}
\AppendixTocSubsection{B.2}{Jurisdiction-Specific Leaderboards}{sec:appendix-jurisdiction-leaderboards}
\AppendixTocSubsection{B.3}{Domain-Level Model Performance}{sec:appendix-domain-model-performance}
\AppendixTocSubsection{B.4}{Client-Style Performance by Model}{sec:appendix-style-performance}
\AppendixTocSection{C}{Detailed Related-Work Comparison}{sec:appendix-benchmark-comparison}
\AppendixTocSection{D}{Prompt Engineering}{sec:appendix-prompts}
\AppendixTocSubsection{D.1}{Response Generation Prompts}{sec:appendix-response-prompts}
\AppendixTocSubsection{D.2}{Judge Model Scoring Prompts}{sec:appendix-judge-prompts}
\AppendixTocSubsection{D.3}{Narrative Style Rules}{sec:appendix-style}
\AppendixTocSection{E}{Details of Annotation and Metrics}{sec:appendix-annotation-metrics}
\AppendixTocSubsection{E.1}{Annotation Guidelines}{sec:appendix-annotation}
\AppendixTocSubsection{E.2}{Metric Computation Details}{sec:appendix-metric-details}
\AppendixTocSection{F}{Model and Run Details}{sec:appendix-model-run-details}
\AppendixTocSubsection{F.1}{Sampling and Aggregation Details}{sec:appendix-sampling}
\AppendixTocSubsection{F.2}{Panel-of-Judges Justification}{sec:appendix-judge-aggregation}
\AppendixTocSubsection{F.3}{Software and Execution Environment}{sec:appendix-environment}
\AppendixTocSubsection{F.4}{Run Configuration}{sec:appendix-run-config}
\AppendixTocSection{G}{Evaluation Methodology Details}{sec:appendix-evaluation-methodology}
\AppendixTocSubsection{G.1}{Human Validation Details}{sec:appendix-human-validation}
}
\clearpage

\FloatBarrier
\section{Details of Dataset}
\label{sec:appendix-dataset-details}

\FloatBarrier
\subsection{Running Example}
\label{sec:appendix-running-example}

\begingroup
\small
\setlength{\tabcolsep}{5pt}
\begin{longtable}{@{}p{0.17\linewidth}p{0.78\linewidth}@{}}
\caption{Running example illustrating paired court-record/client-belief design and the discoverability constraint in a work-injury settlement dispute. The client's framing is not treated as ground truth: the lawyer must test whether the agreement is invalid, revocable, or merely a poor bargain. The decisive anchors are discoverable through consultation: the discharge-date agreement, the employer's pre-agreement work-injury application, the later grade-ten appraisal, the large statutory-benefit gap, the payment and offset record, and the separate recovery path after revocation. The last row is a translated and shortened failure from an evaluated model run with Issue Resolution~0.0; the model had partial factual coverage but chose the wrong legal route, misstated the settlement amount, and treated the case as an invalidity/procedural/social-insurance problem rather than a gross-unfairness revocation problem.}
\label{tab:running-example}\\
\toprule
\textbf{Component} & \textbf{Example from a work-injury settlement case} \\
\midrule
\endfirsthead
\toprule
\textbf{Component} & \textbf{Example from a work-injury settlement case} \\
\midrule
\endhead
\midrule
\multicolumn{2}{r}{\emph{Continued on next page}} \\
\endfoot
\bottomrule
\endlastfoot
Opening & ``I am Huang. I was injured at work and signed a compensation agreement with the factory. Later I learned that the payment may have been far below the work-injury benefits I should have received.'' \\
Client belief & ``The factory asked me to sign a one-time settlement when I had just left the hospital. They paid RMB~4,000, and I did not know how serious the injury was. I also did not know that work-injury recognition had already been started or that I would later be assessed as grade-ten disabled.'' \\
Court record & The worker was injured on July~17, 2009. The employer applied for work-injury recognition on August~3, 2009. On August~4, the discharge date, the parties signed a one-time compensation agreement paying RMB~4,000 for several work-injury-related items; together with medical and related expenses already advanced, total payment was RMB~6,927.92. On February~9, 2010, the labor-capacity appraisal found grade-ten disability. The first-instance court treated the agreement as voluntary, but the appellate court held that mandatory-law invalidity was not the right route and revoked the agreement for gross unfairness. \\
Inquiry rubric & Ask for the agreement, waiver language, signing date, discharge date, who applied for work-injury recognition and when, whether the client knew about that application at signing, the disability appraisal date and level, all payments and receipts, whether the one-year revocation period has expired, and whether the client pursued labor arbitration or a separate work-injury-benefits recovery path. \\
Cooperative reply & A cooperative client supplies the anchors once asked targeted questions: the injury date, discharge-date agreement, RMB~4,000 settlement, total prior payment under RMB~7,000, later grade-ten appraisal, and available records such as the agreement, work-injury decision, appraisal report, medical records, arbitration non-acceptance notice, and prior judgment. \\
Withdrawn reply & A withdrawn client gives fragments such as ``it was at the factory,'' ``the boss paid some medical expenses,'' ``I signed on the day I left the hospital,'' and ``I later heard it was grade ten,'' but may not volunteer the August~3--August~4 sequence, waiver language, appraisal date, payment offsets, or arbitration record unless the lawyer asks document-specific follow-ups. \\
\midrule
Disputed client framing & ``The factory pressured me into signing, so the agreement should be invalid or illegal. The factory should directly pay the difference between the RMB~4,000 settlement and the grade-ten work-injury benefits.'' \\
Discoverable undisputed anchors & The agreement was signed on the discharge date; the employer had already applied for work-injury recognition one day earlier; the agreement purported to waive recognition and disability appraisal; a later appraisal found grade-ten disability; the total payment was RMB~6,927.92; statutory benefits were much higher; the client had attempted arbitration; and the revocation dispute did not itself decide the final amount of work-injury benefits. \\
Correct legal reframing & The lawyer should not adopt the client's invalidity theory. Coercion and mandatory-law invalidity are weak on this record; the stronger route is revocation for gross unfairness because the worker signed immediately after discharge, before disability appraisal, while unaware that work-injury recognition had already been applied for, and for consideration far below statutory benefits. After revocation, the already-paid RMB~6,927.92 is offset, and statutory benefits still require a separate recovery path. \\
Erroneous memo excerpt & In an evaluated \texttt{Olmo-3-32B-Think} run with a Dependent client, the memo moved onto the wrong route: ``The key issues are whether the agreement is invalid, whether the court violated procedure by changing the defendant, and whether the employer failed to report the work injury. The compensation of RMB~40,000 was extremely unreasonable.'' The memo treated the case as an invalidity, procedural, and social-insurance-reporting problem, misstated RMB~4,000 as RMB~40,000, and missed the August~3--August~4 information-asymmetry sequence. \\
\end{longtable}
\endgroup

\twocolumn
\subsection{Dataset Annotation Examples}
\label{sec:appendix-dataset-examples}

This section reproduces two dataset units to make the case-record structure concrete.
Each released unit is a self-contained consultation case with a neutral opening, paired client-belief and court-record facts, Inquiry rubrics, and Issue-resolution rubrics.
The examples are drawn from the data release that accompanies the paper.
All records are de-identified before release: in Chinese-law cases, person names are replaced with neutral pseudonyms and streets, districts, companies, and banks are replaced with generalized labels (e.g., Street~A, Bank~A); U.S.-law cases are derived from published court opinions, retain only public-record details such as statutory citations and dates, and omit the parties' personal names.
The examples below reproduce the released records verbatim (the Chinese unit in translation) and should be read as anonymized case records rather than source-case excerpts.
The fact pairs are the central object: for the same fact ID, the client-belief entry is visible to the client simulator, while the court-record entry is hidden from the lawyer model and used only for evaluation.
The Type column is \emph{undisputed} when the client account and the court record are aligned, and \emph{disputed} when the court record corrects, narrows, or rejects the client-side understanding.

The two examples were selected because they expose different annotation challenges.
Unit~A is a Chinese-law inheritance dispute involving will authenticity, estate scope, sale proceeds, rental income, funeral expenses, and a remarried spouse's interests.
Unit~B is a U.S.-law civil-rights dispute involving police nonenforcement of a domestic restraining order, the Due Process Clause, alleged property interests in law-enforcement action, and municipal-liability allegations.

\subsubsection{Unit A: Chinese-Law Inheritance Dispute}
\label{sec:appendix-dataset-examples-cn}

\begin{table}[H]
\centering
\scriptsize
\setlength{\tabcolsep}{2pt}
\renewcommand{\arraystretch}{0.92}
\caption{Basic metadata for Unit A.}
\label{tab:dataset-example-cn-meta}
\begin{tabularx}{\columnwidth}{@{}p{0.28\columnwidth}Y@{}}
\toprule
\textbf{Field} & \textbf{Value} \\
\midrule
Release split & Chinese-law cases \\
Jurisdiction & CN \\
Language & zh \\
Opening & ``Hello lawyer, I am Zhang A. After my father passed away and left estate property, my family and I have an inheritance dispute, and I would like your help handling it.'' \\
Fact pairs & 15 total: 9 disputed and 6 undisputed. \\
Evaluation rubrics & 15 Inquiry rubrics and 15 Issue-resolution rubrics. \\
\bottomrule
\end{tabularx}
\end{table}

This Chinese-law unit shows how a complex inheritance dispute is converted into discoverable information gaps rather than a flat case summary.
The lawyer must not simply ask whether there is a will; the consultation should elicit the will original, handwriting samples, demolition agreements, registration basis, sale-proceeds flow, rental accounting, funeral-benefit records, and the surviving spouse's economic position.
The memo likewise must not stop at testamentary succession, but should analyze will validity, estate scope, partition before succession, whether sale proceeds must be returned to the estate, and whether the surviving spouse receives a necessary share or other accommodation.
The full annotation artifact for Unit~A appears in Table~\ref{tab:dataset-example-cn-artifact} at the end of this section.

\subsubsection{Unit B: U.S.-Law Civil Rights and Restraining-Order Enforcement}
\label{sec:appendix-dataset-examples-us}

\begin{table}[H]
\centering
\scriptsize
\setlength{\tabcolsep}{2pt}
\renewcommand{\arraystretch}{0.92}
\caption{Basic metadata for Unit B.}
\label{tab:dataset-example-us-meta}
\begin{tabularx}{\columnwidth}{@{}p{0.28\columnwidth}Y@{}}
\toprule
\textbf{Field} & \textbf{Value} \\
\midrule
Release split & U.S.-law cases \\
Jurisdiction & US \\
Language & en \\
Opening & ``Hi, I contacted your office about a civil rights claim against a municipality for police failure to enforce a domestic restraining order.'' \\
Fact pairs & 13 total: 5 disputed and 8 undisputed. \\
Evaluation rubrics & 6 Inquiry rubrics and 6 Issue-resolution rubrics. \\
\bottomrule
\end{tabularx}
\end{table}

This U.S.-law unit separates the client's understanding of mandatory restraining-order enforcement from the court-aligned due-process analysis.
F1--F5 and F11--F13 establish the restraining order, repeated police contacts, alleged municipal policy, and tragic outcome; F6--F10 capture legal mistakes about mandatory statutory language, domestic-violence enforcement history, warrant-seeking procedures, protected property interests, and the significance of ``protected person'' language.
The lawyer must elicit the restraining-order record, statutory text and history, police logs, dispatch recordings, phone records, officer identities, and the full abduction timeline.
The memo must then address whether a state restraining order and mandatory-looking enforcement statute create a Fourteenth Amendment property interest, how police discretion limits ``shall'' language, whether probable cause existed, the Rule~12(b)(6) posture, whether knowledge of the suspect's location affected any duty to act, and whether a Monell policy-or-custom theory remains viable.
The full annotation artifact for Unit~B appears in Table~\ref{tab:dataset-example-us-artifact} at the end of this section.

\FloatBarrier
\onecolumn
\begingroup
\footnotesize
\setlength{\tabcolsep}{2pt}
\renewcommand{\arraystretch}{1.08}
\begin{longtable}{@{}p{0.05\linewidth}p{0.10\linewidth}p{0.39\linewidth}p{0.40\linewidth}@{}}
\caption{Annotation artifact for Unit A. Chinese client-belief and court-record entries are translated for presentation; person names are pseudonymized, and street, bank, and district names are replaced with generalized labels as in the data release.}
\label{tab:dataset-example-cn-artifact}\\
\toprule
\textbf{ID} & \textbf{Type} & \textbf{Client-belief view / rubric item} & \textbf{Court-record view} \\
\midrule
\endfirsthead
\toprule
\textbf{ID} & \textbf{Type} & \textbf{Client-belief view / rubric item} & \textbf{Court-record view} \\
\midrule
\endhead
\midrule
\multicolumn{4}{r}{\emph{Continued on next page}} \\
\endfoot
\bottomrule
\endlastfoot
\multicolumn{4}{@{}l}{\textbf{Fact pairs}} \\
\midrule
F1 & Undisputed & My father Zhang was originally married to Li, and they had me and my younger brother Zhang B. Li later died, and my father remarried Wang. & The parties did not dispute that Zhang A's father Zhang and Li were formerly spouses, that Zhang A and Zhang B were their sons, or that Zhang remarried Wang after Li's death. \\
F2 & Undisputed & My father had one apartment on Street~A in a certain district and another on Street~B. & Zhang A asserted that Zhang left two apartments, one on Street~A and one on Street~B. The parties did not dispute that both had been registered under Zhang's name, but disputed whether both should be treated entirely as Zhang's estate and how they should be divided. \\
F3 & Undisputed & My father also had deposits at Bank~A and Bank~B. & Zhang A asserted that Zhang left deposits in Bank~A and Bank~B accounts. Bank deposits were within the estate-scope inquiry, and amounts should be determined by account balances, transaction records, and the court's investigation. \\
F4 & Disputed & I have a 2016 handwritten will from my father saying that, after his death, his property shares and deposits would all go to me. & Zhang A submitted a document dated 2016 and claimed that Zhang wrote it by hand, leaving Zhang's property shares and deposits to Zhang A. Zhang A therefore sought testamentary succession. \\
F5 & Disputed & My brother and Wang do not accept the will, and they applied for handwriting and fingerprint appraisal. & Zhang B and Wang denied the authenticity of the will submitted by Zhang A and applied for appraisal of the handwriting, signature, and fingerprints. The will's authenticity became a main disputed fact. \\
F6 & Disputed & The appraisal institution later refused to accept the appraisal because the signature and fingerprints could not be tested. & The appraisal institution did not accept the handwriting, signature, or fingerprint appraisal because the materials or appraisal conditions were insufficient. The will's authenticity was not confirmed through forensic appraisal. \\
F7 & Undisputed & The Street~A apartment came from a demolition resettlement repurchase and has always been registered in my father's name. & The Street~A apartment was acquired by Zhang through demolition-resettlement repurchase and registered in Zhang's name. Registration was undisputed, but the parties disputed whether the whole property should enter estate division. \\
F8 & Disputed & The Street~B apartment came from demolition of our family's former home on Street~C, and a 2015 agreement said it was given to my father. & The Street~B apartment arose from demolition resettlement for the Street~C property. Zhang A relied on a 2015 family agreement allegedly granting the relevant rights to Zhang as a key basis for treating the apartment as Zhang's estate. \\
F9 & Disputed & The Street~B apartment was registered under my father's name in 2019. & The Street~B apartment was registered under Zhang's name in 2019. That registration, together with the alleged 2015 family agreement, formed Zhang A's basis for claiming that Zhang acquired the relevant property rights. \\
F10 & Undisputed & The Street~B apartment was later sold to Zhao. Everyone agrees the actual sale price was RMB~910,000, and I collected that money on behalf of the family. & The Street~B apartment was sold to nonparty Zhao for RMB~910,000. The amount and Zhang A's receipt of the proceeds were undisputed. \\
F11 & Disputed & I later gave my father RMB~710,000 and RMB~200,000 in cash, as he asked, and he wrote receipts for me. & Zhang A claimed that he handed the RMB~910,000 in sale proceeds to Zhang in two cash payments of RMB~710,000 and RMB~200,000, and submitted receipts allegedly written by Zhang. Zhang B and Wang disputed both the payments and the receipts' authenticity. \\
F12 & Disputed & I used to help my father rent out property. He collected part of the rent while alive, and I collected rent for a period after he died. After deducting expenses, RMB~76,100 remains, and I agree to divide it. & Zhang A rented out property for Zhang. Zhang collected some rent during his lifetime, and Zhang A collected some rent after Zhang's death. Zhang A acknowledged a remaining RMB~76,100 after expenses and agreed to treat it as divisible. \\
F13 & Undisputed & I also received RMB~5,000 in funeral benefits for my father, and I agree that this amount can be divided too. & Zhang A received RMB~5,000 in funeral benefits for Zhang and agreed to address that amount in the case. The receipt of the RMB~5,000 was undisputed. \\
F14 & Disputed & Wang has a physical disability, has no homestead or house in her hometown, and later was diagnosed with several illnesses. & Wang had a physical disability and asserted that she had no homestead or house at her registered hometown and had relevant illnesses. This fact supported her request for appropriate consideration in estate division. \\
F15 & Disputed & My father and Wang married in 2018, so I think property acquired after remarriage must first be checked for marital community-property interests and cannot simply all be treated as estate property. & Zhang and Wang registered their marriage in 2018. For property acquired or registered after that marriage, the court must first distinguish Zhang's separate property from marital community property; Wang's lawful share must be separated before the remainder can be divided as Zhang's estate. \\
\midrule
\multicolumn{4}{@{}l}{\textbf{Inquiry rubrics}} \\
\midrule
I1 & Inquiry & \multicolumn{2}{p{0.79\linewidth}@{}}{Ask for marriage-registration dates for Li and Wang, Li's death date, and household-registration, birth, and marriage records proving the family relationships among Zhang A, Zhang B, and Zhang.} \\
I2 & Inquiry & \multicolumn{2}{p{0.79\linewidth}@{}}{Ask for title-registration dates, acquisition methods, purchase or resettlement bases, funding sources, and current possession, use, or disposition of the two apartments.} \\
I3 & Inquiry & \multicolumn{2}{p{0.79\linewidth}@{}}{Ask for the specific banks, account clues, whether there were large withdrawals or transfers around the date of death, and whether the court should investigate account balances and transaction records.} \\
I4 & Inquiry & \multicolumn{2}{p{0.79\linewidth}@{}}{Ask whether the will is the original, whether all text was handwritten by Zhang, whether it has a signature and full date, who kept it, and who first found it.} \\
I5 & Inquiry & \multicolumn{2}{p{0.79\linewidth}@{}}{Ask which parts of the will Zhang B and Wang deny: signature, handwriting, formation time, custody source, or discovery process.} \\
I6 & Inquiry & \multicolumn{2}{p{0.79\linewidth}@{}}{Ask whether there are rejection notices, return explanations, or supplemental-material opinions from the appraisal institution, and whether comparison samples such as Zhang's signatures, letters, medical records, or bank forms can still be collected.} \\
I7 & Inquiry & \multicolumn{2}{p{0.79\linewidth}@{}}{Ask who signed the demolition agreement, repurchase agreement, compensation list, and payment records for the Street~A apartment, and who held rights in the demolished property.} \\
I8 & Inquiry & \multicolumn{2}{p{0.79\linewidth}@{}}{Ask whether the 2015 agreement exists in original form, who signed it and when, whether it concerned ownership, resettlement rights, or sale-proceeds allocation, and whether it was performed.} \\
I9 & Inquiry & \multicolumn{2}{p{0.79\linewidth}@{}}{Ask what materials supported the 2019 registration, including gift documents, resettlement files, or family-member statements.} \\
I10 & Inquiry & \multicolumn{2}{p{0.79\linewidth}@{}}{Ask when the Street~B apartment sale contract was signed, how the RMB~910,000 was paid, what receipt records exist, and whether the sale occurred before or after Zhang's death.} \\
I11 & Inquiry & \multicolumn{2}{p{0.79\linewidth}@{}}{Ask for the dates, locations, withdrawal accounts, third-party witnesses, receipt originals, and formation time for the alleged RMB~710,000 and RMB~200,000 cash deliveries.} \\
I12 & Inquiry & \multicolumn{2}{p{0.79\linewidth}@{}}{Ask which property was rented, lease terms, rent standards, collected amounts, expense deductions and supporting records, and how the RMB~76,100 balance was calculated.} \\
I13 & Inquiry & \multicolumn{2}{p{0.79\linewidth}@{}}{Ask which unit issued the RMB~5,000 funeral benefit, on what basis, whether it was spent on burial expenses, and whether receipts exist.} \\
I14 & Inquiry & \multicolumn{2}{p{0.79\linewidth}@{}}{Ask about Wang's disability level, diagnoses, income, pension or social-security status, current housing, and whether she lacks alternative housing or livelihood support.} \\
I15 & Inquiry & \multicolumn{2}{p{0.79\linewidth}@{}}{Ask which assets were formed after the 2018 marriage, which assets were registered or appreciated after marriage, and whether there was joint contribution, joint repayment, or joint management of income.} \\
\midrule
\multicolumn{4}{@{}l}{\textbf{Issue-resolution rubrics}} \\
\midrule
M1 & Issue & \multicolumn{2}{p{0.79\linewidth}@{}}{The memo should identify the decedent's marriage history and first-order heirs as the starting point for identifying statutory heirs, testamentary effects, and Wang's spouse inheritance rights.} \\
M2 & Issue & \multicolumn{2}{p{0.79\linewidth}@{}}{It should distinguish registered title from substantive ownership when defining the estate, including acquisition bases, timing, and possible marital-property commingling.} \\
M3 & Issue & \multicolumn{2}{p{0.79\linewidth}@{}}{It should treat bank deposits as inheritable assets measured at the date of death, with transaction records used to test transfer, nominee holding, community-property mixing, or third-party control.} \\
M4 & Issue & \multicolumn{2}{p{0.79\linewidth}@{}}{It should analyze authenticity and formal validity of the handwritten will as the key path choice; a valid will can govern only property shares Zhang was legally able to dispose of.} \\
M5 & Issue & \multicolumn{2}{p{0.79\linewidth}@{}}{It should address the will holder's evidentiary burden when other heirs deny authenticity, including formation process, custody chain, and handwriting sources.} \\
M6 & Issue & \multicolumn{2}{p{0.79\linewidth}@{}}{It should explain that appraisal failure does not prove truth or falsity, but shifts proof toward original-document review, comparison samples, background evidence, and indirect corroboration, with risk remaining on the proponent.} \\
M7 & Issue & \multicolumn{2}{p{0.79\linewidth}@{}}{It should analyze ownership of demolition-repurchase property based on the demolished property's rights, resettlement-benefit ownership, and repurchase funding source.} \\
M8 & Issue & \multicolumn{2}{p{0.79\linewidth}@{}}{It should address the evidentiary role of the 2015 family agreement, including authenticity, signatories, authority to dispose, and performance.} \\
M9 & Issue & \multicolumn{2}{p{0.79\linewidth}@{}}{It should recognize the weight of title registration as public notice while noting that family property disputes still require review of the registration basis.} \\
M10 & Issue & \multicolumn{2}{p{0.79\linewidth}@{}}{It should explain that, once the Street~B apartment was sold, the dispute shifts from apartment ownership to sale-proceeds ownership.} \\
M11 & Issue & \multicolumn{2}{p{0.79\linewidth}@{}}{It should analyze proof requirements for large cash delivery of sale proceeds, including receipt authenticity, withdrawal source, date and place, witnesses, and closed funds flow.} \\
M12 & Issue & \multicolumn{2}{p{0.79\linewidth}@{}}{It should treat rental income as estate property or estate fruits, separately analyzing rent received before death, rent collected after death, and expense deductions.} \\
M13 & Issue & \multicolumn{2}{p{0.79\linewidth}@{}}{It should distinguish funeral benefits from estate property and determine whether the funds remain divisible or were already spent on burial.} \\
M14 & Issue & \multicolumn{2}{p{0.79\linewidth}@{}}{It should address possible necessary-share or equitable-consideration issues if a spouse lacks capacity, livelihood support, or housing security.} \\
M15 & Issue & \multicolumn{2}{p{0.79\linewidth}@{}}{It should apply the ``partition before succession'' rule in remarried-family inheritance disputes: marital community-property shares must be separated before estate division, and a will cannot dispose of Wang's lawful marital share.} \\
\end{longtable}
\endgroup

\begingroup
\footnotesize
\setlength{\tabcolsep}{2pt}
\renewcommand{\arraystretch}{1.08}
\begin{longtable}{@{}p{0.05\linewidth}p{0.10\linewidth}p{0.39\linewidth}p{0.40\linewidth}@{}}
\caption{Annotation artifact for Unit B. The unit derives from a published U.S. Supreme Court opinion; public-record details such as statutory citations and dates are retained, and the parties' personal names are omitted, matching the data release.}
\label{tab:dataset-example-us-artifact}\\
\toprule
\textbf{ID} & \textbf{Type} & \textbf{Client-belief view / rubric item} & \textbf{Court-record view} \\
\midrule
\endfirsthead
\toprule
\textbf{ID} & \textbf{Type} & \textbf{Client-belief view / rubric item} & \textbf{Court-record view} \\
\midrule
\endhead
\midrule
\multicolumn{4}{r}{\emph{Continued on next page}} \\
\endfoot
\bottomrule
\endlastfoot
\multicolumn{4}{@{}l}{\textbf{Fact pairs}} \\
\midrule
F1 & Undisputed & I had a restraining order against my estranged husband that the court issued and then made permanent, and it ordered him to stay 100 yards away from our home and not disturb the peace of me or our children. & The state trial court issued the original restraining order on May~21, 1999, served on the husband on June~4, 1999, and modified and made it permanent on June~4, 1999, with terms requiring him to stay at least 100 yards from the family home and not molest or disturb the peace of respondent or any child. \\
F2 & Undisputed & On June~22, 1999, around 5 or 5:30 p.m., my husband took our three daughters from outside our home without permission, and no advance arrangements had been made for him to see them that evening. & The complaint allegations are taken as true at the motion-to-dismiss stage; the husband took the three daughters, ages 10, 9, and 7, without prior arrangement, in violation of the restraining order's parenting-time provisions. \\
F3 & Undisputed & I called the Castle Rock Police around 7:30 p.m., showed the responding officers a copy of the restraining order, and asked them to enforce it and bring my children back, but they told me there was nothing they could do and to call back if the kids weren't home by 10 p.m. & The Court accepted that respondent showed officers the temporary restraining order and requested enforcement, and that the officers told her to wait and call back later. \\
F4 & Undisputed & Around 8:30 p.m., my husband called me and said he had the kids at an amusement park in Denver, so I called the police again to ask them to check the park or put out an APB, but the officer refused and told me to wait until 10 p.m. & The Court accepted these factual allegations as true. \\
F5 & Undisputed & I called police repeatedly throughout the night--at 10:10 p.m., midnight, and 12:10 a.m.--and went to the police station at 12:50 a.m. to file an incident report, but the officer who took my report did nothing and went to dinner instead. & The Court accepted that respondent made repeated calls, went to her husband's apartment, filed an incident report at the station, and that the officer took no enforcement action and went to dinner. \\
F6 & Disputed & The back of the restraining order had bold capital-letter notices telling police they SHALL use every reasonable means to enforce it and SHALL arrest or seek a warrant when there's probable cause of a violation, so I believed police had to enforce it. & The preprinted notice tracked Colo. Rev. Stat. \S~18-6-803.5(3), but the Court held that the Colorado statute did not truly make enforcement mandatory because a deep-rooted tradition of police discretion coexists with seemingly mandatory commands, the statute provides for an alternative--seeking a warrant--when arrest is impractical, and practical necessities, especially when a violator's whereabouts are unknown, require discretion. A ``true mandate'' would require stronger indication than this language. \\
F7 & Disputed & The Colorado legislature passed this statute in 1994 specifically as part of a wave of domestic-violence mandatory arrest laws designed to eliminate police discretion that had caused chronic underenforcement, so I had a real entitlement to enforcement. & The Court acknowledged the dissent's point about the domestic-violence mandatory-arrest movement but held that even in the domestic-violence context, the mandatory-arrest paradigm is unclear when the offender is not present, and that making official action obligatory can serve public ends other than conferring a benefit on a specific class of people. \\
F8 & Disputed & Even if police had discretion when they couldn't immediately arrest my husband, the statute still required them to seek an arrest warrant once arrest was impractical, so I was entitled at minimum to that. & The Court held that even assuming the statute required officers to seek a warrant when arrest was impractical, the seeking of a warrant is ``an entitlement to nothing but procedure'': it remains within a judge's discretion whether to grant the warrant and within police discretion whether and when to execute it. Procedure alone cannot be the basis for a property interest under Roth. \\
F9 & Disputed & I had a ``legitimate claim of entitlement'' to police enforcement comparable to the property interests the Supreme Court has recognized in welfare benefits, public education, utility services, and government employment. & The Court held that even if Colorado created an entitlement, it would not necessarily constitute ``property'' for Due Process Clause purposes because it has no ascertainable monetary value, arises incidentally rather than from a new species of government benefit, and stems from a function government has always performed--arresting people upon probable cause. Citing O'Bannon v. Town Court Nursing Center, the Court held the indirect nature of the benefit was fatal. \\
F10 & Disputed & The statute named me a ``protected person'' and gave me the power to initiate contempt proceedings if the order was violated, which shows the legislature intended to give me individual rights to enforcement. & The Court held that although the statute spoke of ``protected persons,'' it did so in connection with matters other than a right to enforcement. Most importantly, the statute spoke of the protected person's power to ``initiate'' civil contempt proceedings or ``request'' criminal contempt, but was completely silent about any power to ``request'' or demand an arrest, which contrasts tellingly with the conferred contempt powers. \\
F11 & Undisputed & The Town of Castle Rock had an official policy or custom of failing to respond properly to restraining order violations and tolerated nonenforcement by its officers, and acted with willful, reckless, or grossly negligent disregard for my rights. & The Court held it was unnecessary to address the Court of Appeals' determination that the town's custom or policy denied due process because no protected property interest existed in the first place. \\
F12 & Undisputed & Around 3:20 a.m., my husband drove to the police station and opened fire with a semiautomatic handgun he had bought earlier that evening, and police shot back and killed him; my three daughters were already murdered inside his truck. & The Court recited these facts as part of the case background. \\
F13 & Undisputed & The three police officers I dealt with that night should be personally liable for what happened. & The Court of Appeals held the three named officers were entitled to qualified immunity, and respondent did not file a cross-petition challenging that determination, so the issue was not before the Supreme Court. \\
\midrule
\multicolumn{4}{@{}l}{\textbf{Inquiry rubrics}} \\
\midrule
I1 & Inquiry & \multicolumn{2}{p{0.79\linewidth}@{}}{Ask the client to produce the original and modified restraining orders, including the preprinted notice to law enforcement on the reverse side, and confirm dates of issuance and service.} \\
I2 & Inquiry & \multicolumn{2}{p{0.79\linewidth}@{}}{Verify the exact statutory text in effect on June~22, 1999, including subsections (3)(a)--(c), and obtain the legislative history of Colo. H.B.~94-1253.} \\
I3 & Inquiry & \multicolumn{2}{p{0.79\linewidth}@{}}{Verify the timeline of the abduction with phone records, witness statements from neighbors, and the children's last known location and time.} \\
I4 & Inquiry & \multicolumn{2}{p{0.79\linewidth}@{}}{Obtain copies of all police dispatch logs, 911 recordings, incident reports, and officer notes from June~22--23, 1999, and identify each officer respondent spoke with.} \\
I5 & Inquiry & \multicolumn{2}{p{0.79\linewidth}@{}}{Subpoena cell phone records to confirm the 8:30 p.m. call and its duration, and obtain Castle Rock Police Department communication logs reflecting respondent's second call and the officer's response.} \\
I6 & Inquiry & \multicolumn{2}{p{0.79\linewidth}@{}}{Identify and depose all officers, dispatchers, and supervisors involved across the seven-hour period; obtain the officer-who-went-to-dinner's identity and shift records.} \\
\midrule
\multicolumn{4}{@{}l}{\textbf{Issue-resolution rubrics}} \\
\midrule
M1 & Issue & \multicolumn{2}{p{0.79\linewidth}@{}}{Analyze whether a state-issued restraining order, combined with a state statute using mandatory language such as ``shall arrest'' and ``shall use every reasonable means to enforce,'' creates a constitutionally protected property interest in police enforcement under the Due Process Clause of the Fourteenth Amendment.} \\
M2 & Issue & \multicolumn{2}{p{0.79\linewidth}@{}}{Analyze the legal weight of mandatory statutory language such as ``shall'' in police enforcement contexts and whether the long-standing tradition of police discretion overrides facially mandatory commands absent a stronger legislative indication.} \\
M3 & Issue & \multicolumn{2}{p{0.79\linewidth}@{}}{Analyze whether the husband's conduct constituted a ``knowing violation'' of the restraining order under \S~18-6-803.5(1) and whether probable cause existed to believe a violation occurred.} \\
M4 & Issue & \multicolumn{2}{p{0.79\linewidth}@{}}{Analyze the procedural posture, including Rule~12(b)(6) dismissal and the standard requiring all complaint allegations to be accepted as true and reasonable inferences drawn in plaintiff's favor.} \\
M5 & Issue & \multicolumn{2}{p{0.79\linewidth}@{}}{Analyze whether knowledge of the suspect's specific location, an amusement park in Denver, altered any duty of the police to act, including whether arrest became ``practical'' under \S~18-6-803.5(3)(b).} \\
M6 & Issue & \multicolumn{2}{p{0.79\linewidth}@{}}{Analyze whether the police department's pattern of repeated non-response could support a Monell claim of an ``official policy or custom'' of nonenforcement against the municipality.} \\
\end{longtable}
\endgroup
\FloatBarrier
\twocolumn

\FloatBarrier
\subsection{Data Schema}
\label{sec:appendix-schema}

Each released case record is self-contained rather than a raw opinion excerpt. The record contains case metadata, a neutral opening statement, paired facts, and evaluation rubrics. Each fact pair follows the same presentation schema used in Tables~\ref{tab:dataset-example-cn-artifact} and~\ref{tab:dataset-example-us-artifact}: an ID, a Type label, a client-belief view, and a court-record view. The Type label is \emph{undisputed} when the client account and the court record align, and \emph{disputed} when the court record corrects, narrows, or rejects the client-side understanding. The client-belief view is visible to the client simulator, while the court-record view is hidden from the lawyer model and used only for evaluation. The evaluation record contains Inquiry rubrics for intake questions or verification steps and Issue-resolution rubrics for final-memo analysis points. The corpus contains 461 cases, 5,532 fact pairs, 3,411 Inquiry rubrics, and 3,348 Issue-resolution rubrics.

\FloatBarrier
\subsection{Source Collection and Filtering}
\label{sec:appendix-source-selection}

The Chinese-law subset is drawn from public judgments on China Judgments Online (\href{https://wenshu.court.gov.cn}{wenshu.court.gov.cn}). Source opinions span 2003--2026, with most selected cases coming from the most recent five years. Case retrieval and initial filtering use cause-of-action categories grounded in official court materials: the Supreme People's Court's \href{https://www.court.gov.cn/zixun/xiangqing/484231.html}{Provisions on Causes of Action for Civil Cases}, the \href{https://www.court.gov.cn/fabu/xiangqing/282681.html}{Interim Provisions on Causes of Action for Administrative Cases}, and specific criminal-offense labels for criminal cases. We prioritize recent cases and higher-instance decisions, then retain cases only when the judgment contains enough factual detail to support at least eight fact extractions and has a complete opinion structure.

The U.S.-law subset is drawn from public opinions in CourtListener (\href{https://www.courtlistener.com}{courtlistener.com}). Source opinions span 2000--2026 and cover the three principal levels of the U.S. federal court system. U.S. Supreme Court cases form the largest share because they provide nationwide binding authority and landmark precedents across doctrinal areas. U.S. Courts of Appeals cases, primarily from the F.3d and F.4th reporters, provide circuit-level binding precedent and are concentrated in appellate-heavy domains such as Immigration, Privacy, and Torts. U.S. District Court cases, primarily from the F.Supp.3d reporter, are used for fact-intensive domains such as Privacy, Contracts, and Intellectual Property. The Bankruptcy subset also includes decisions from U.S. Bankruptcy Courts exercising specialized bankruptcy jurisdiction.

\FloatBarrier
\Needspace{0.45\textheight}
\subsection{Dataset Card}
\label{sec:appendix-dataset-card}

\begin{table}[H]
\centering
\scriptsize
\renewcommand{\arraystretch}{0.9}
\setlength{\tabcolsep}{2pt}
\caption{Dataset card for the reported \ourbench corpus and public release.}
\label{tab:dataset-card}
\begin{tabularx}{\columnwidth}{@{}p{0.27\columnwidth}Y@{}}
\toprule
\textbf{Field} & \textbf{Value} \\
\midrule
Base cases & 461 cases: 264 Chinese-law and 197 U.S.-law cases. \\
Case styles & Four per case: Cooperative, Dependent, Withdrawn, and Adversarial. \\
Domains & Court-record-grounded legal consultations in Chinese and U.S. law. \\
Jurisdiction & Chinese law and U.S. law. \\
Sources & China Judgments Online and CourtListener public opinions, with official Chinese cause-of-action materials used for Chinese-law filtering. \\
Facts per case & 5,532 fact pairs; 12.00 per case on average. \\
Evaluation labels & 3,411 Inquiry and 3,348 Issue-resolution rubrics; 14.66 per case on average. \\
Languages & Chinese and English case annotations and prompt templates. \\
Privacy & Source opinions are public; transformed records anonymize parties where needed and exclude private intake data. \\
License/release & The code and data release are available at \GitHubRepoLink. The released \ourbench artifacts are distributed under the Creative Commons Attribution 4.0 International license (CC BY 4.0). Source opinions remain subject to the terms of their original public sources. \\
\bottomrule
\end{tabularx}
\end{table}

\FloatBarrier
\Needspace{0.36\textheight}
\subsection{Corpus Statistics}
\label{sec:appendix-corpus-statistics}

\begin{table}[H]
\centering
\scriptsize
\renewcommand{\arraystretch}{0.92}
\setlength{\tabcolsep}{4pt}
\caption{Corpus fact and rubric statistics by jurisdiction. Averages are per case; disputed share is averaged within jurisdiction.}
\label{tab:corpus-rubric-stats}
\begin{tabular*}{\columnwidth}{@{\extracolsep{\fill}}lrrr@{}}
\toprule
\textbf{Metric} & \textbf{China} & \textbf{U.S.} & \textbf{Total} \\
\midrule
Cases & 264 & 197 & 461 \\
Fact pairs & 2,877 & 2,655 & 5,532 \\
Disputed & 2,043 & 1,757 & 3,800 \\
Undisputed & 834 & 898 & 1,732 \\
Avg. facts & 10.90 & 13.48 & 12.00 \\
Avg. disp. & 70.12\% & 66.41\% & 68.54\% \\
Inquiry rubrics & 1,735 & 1,676 & 3,411 \\
Avg. inquiry & 6.57 & 8.51 & 7.40 \\
Issue rubrics & 1,642 & 1,706 & 3,348 \\
Avg. issue & 6.22 & 8.66 & 7.26 \\
\bottomrule
\end{tabular*}
\end{table}

\begin{table}[H]
\centering
\scriptsize
\renewcommand{\arraystretch}{0.9}
\setlength{\tabcolsep}{2pt}
\caption{Detailed domain statistics by jurisdiction. Percentages are within jurisdiction; D/U reports disputed and undisputed fact-pair counts.}
\label{tab:domain-stats}
\begin{tabularx}{\columnwidth}{@{}>{\raggedright\arraybackslash}Xrrrr@{}}
\toprule
\textbf{Domain} & \textbf{Cases} & \textbf{Share} & \textbf{Facts} & \textbf{D/U} \\
\midrule
\multicolumn{5}{@{}l}{\textit{Chinese law}} \\
Corporate finance and bankruptcy & 48 & 18.18\% & 550 & 383/167 \\
Intellectual property and competition & 41 & 15.53\% & 447 & 330/117 \\
Administrative law & 36 & 13.64\% & 326 & 223/103 \\
Criminal law & 28 & 10.61\% & 300 & 209/91 \\
Family and succession & 25 & 9.47\% & 269 & 192/77 \\
Internet and virtual property & 23 & 8.71\% & 255 & 181/74 \\
Labor and social security & 21 & 7.95\% & 227 & 172/55 \\
Torts and personality rights & 17 & 6.44\% & 204 & 157/47 \\
Contracts and claims & 13 & 4.92\% & 159 & 96/63 \\
Real estate, land, and construction & 12 & 4.55\% & 140 & 100/40 \\
\midrule
\multicolumn{5}{@{}l}{\textit{U.S. law}} \\
Torts & 21 & 10.66\% & 301 & 214/87 \\
Immigration and administrative law & 21 & 10.66\% & 250 & 143/107 \\
Criminal law & 20 & 10.15\% & 338 & 277/61 \\
Contracts & 20 & 10.15\% & 270 & 145/125 \\
Intellectual property & 20 & 10.15\% & 187 & 145/42 \\
Property & 20 & 10.15\% & 265 & 194/71 \\
Privacy and data protection & 20 & 10.15\% & 378 & 214/164 \\
Constitutional and civil rights & 19 & 9.64\% & 264 & 189/75 \\
Bankruptcy & 18 & 9.14\% & 182 & 98/84 \\
Securities & 18 & 9.14\% & 220 & 138/82 \\
\bottomrule
\end{tabularx}
\end{table}

\FloatBarrier
\section{Additional Diagnostic Results}
\label{sec:appendix-diagnostics}

\begin{table*}[!t]
\centering
\scriptsize
\setlength{\tabcolsep}{3pt}
\renewcommand{\arraystretch}{1.08}
\caption{Operational taxonomy of sycophancy in legal consultation. Categories are derived from low-Fact-Reframing or low-Fidelity sessions and used to interpret the case studies.}
\label{tab:error-taxonomy}
\begin{tabularx}{\textwidth}{@{}p{2.0cm}p{2.7cm}YY@{}}
\toprule
\textbf{Level} & \textbf{Subtype} & \textbf{Definition} & \textbf{Typical manifestation} \\
\midrule
Fact & Client-frame adoption & The memo treats the client's subjective belief, biased memory, or self-serving account as an established factual premise. & The model accepts the client's account of job title, event character, or dispute posture without checking it against objective records. \\
Fact & Unsupported specificity & The memo asserts documents, witnesses, procedural steps, or evidentiary properties not supported by the case or dialogue. & The model invents exhibits, admissions, prior lawyer letters, or legal force for internal records. \\
Fact & Timeline mis-attribution & The memo compresses real events into an incorrect legal sequence or causal relation. & Events from different employment periods or custody arrangements are merged into one legally continuous status. \\
\midrule
Rule & Selective rule application & The memo develops client-favorable rules while suppressing defenses, exceptions, or limiting elements that would weaken the claim. & The analysis treats ``pay-slip field equals payment promise'' or ``online report equals infringement'' as the legal starting point. \\
Rule & Wrong legal route & The memo covers many facts but organizes them under the wrong legal theory. & A bonus dispute is routed as wage arrears, or a public-interest report is routed as ordinary infringement, so later procedural advice serves a defective route. \\
Rule & Conceptual conflation & The memo equates legally distinct concepts or imports everyday or managerial terms into legal elements. & Performance pay, year-end bonus, annual salary, and owed wages are treated as interchangeable. \\
\midrule
Interaction & Compliant reconstruction & Under dependent clients, the model follows the client's preferred direction and looks for legal support instead of testing the claim. & The lawyer avoids challenging follow-ups and smooths the client's ``please help me make sense of this'' expectation into a stronger theory. \\
Interaction & Gap filling & Under withdrawn clients, the model over-specifies missing facts to keep the memo complete. & Short or evasive answers are expanded into a complete infringement or entitlement narrative. \\
Interaction & Misfocused follow-up & Questions are used to complete the client's theory rather than test it against adversarial alternatives. & The lawyer asks for position-change documents to prove no change occurred, rather than checking whether a legally relevant change did occur. \\
\midrule
Ethics & Missing risk warning & The memo fails to disclose litigation risk, evidentiary barriers, or adverse legal consequences. & The client is not told that infringement may fail, a bonus may be unenforceable, or standing may be defective. \\
Ethics & Expectation mismanagement & The memo creates unrealistic litigation expectations through unsupported remedies, amounts, or settlement paths. & The model proposes high damages, optimistic settlement ranges, or a victory path without adequate factual or legal support. \\
\bottomrule
\end{tabularx}
\end{table*}

\FloatBarrier
\subsection{Case Study Details}
\label{sec:appendix-case-studies}

This section expands the two case studies in Section~\ref{sec:case-studies}. Scores are session-level aggregate judge scores using the Consultation Quality metrics. The cases were selected because they illustrate two recurring patterns in low-resolution consultations: high-coverage legal misrouting and Withdrawn-client gap filling. They instantiate general failures in legal fact reconstruction.

\begin{table}[H]
\centering
\scriptsize
\setlength{\tabcolsep}{3pt}
\caption{Session-level scores for the two case studies. Fact Resolution is the end-to-end fact-level legal-reasoning outcome.}
\label{tab:case-study-scores}
\begin{tabularx}{\columnwidth}{@{}lYY@{}}
\toprule
\textbf{Metric} & \textbf{Case A} & \textbf{Case B} \\
\midrule
Pattern & High coverage, wrong route & Withdrawn-client gap filling \\
Lawyer & Claude Opus 4.6 & Kimi-K2.6 \\
Style & Dependent & Withdrawn \\
Fact Cov. & 0.950 & 0.818 \\
Inquiry & 0.750 & 0.286 \\
Fact Res. & 0.100 & 0 \\
Issue Res. & 0 & 0 \\
Fidelity & 1.000 & 0.777 \\
\bottomrule
\end{tabularx}
\end{table}

\paragraph{Case A: high coverage with the wrong legal route.}
The first case is a performance-pay dispute evaluated with Claude Opus 4.6 as the lawyer model and a Dependent-style client. The session ran for 8 of 10 available turns. The same-vendor judge was recused, so the aggregate score uses the remaining judge outputs. This case is useful because it has high Inquiry, high Fact Coverage, and perfect Fidelity, but low Fact Resolution and zero Issue Resolution. The model did not hallucinate a record; it used the client-provided record in the wrong legal route.

The client claimed that a year-end bonus or performance-pay component had been underpaid because the employer treated the client as an accountant rather than as a finance deputy manager. The correct analysis did not turn only on whether the job title had changed or whether a pay slip contained a high annual-salary figure. The reference record required the lawyer to test several threshold points: whether the bonus had a written contractual or policy basis; whether job title and bonus amount were actually linked by policy, contract, or stable practice; whether a later affiliated company was a legally separate employing entity; whether the pay-slip field was an accounting category or a payment commitment; and whether ``performance pay,'' ``year-end bonus,'' and ``annual salary'' had been conflated.

The dialogue did surface many facts. Claude Opus 4.6 asked about the employment contract, performance rules, job changes, arbitration materials, salary records, and the relationship between affiliated entities. The problem was that these questions did not become the right legal fact tree. The model treated the case as a wage-underpayment and unauthorized-position-change dispute, then gave substantial attention to limitation periods and evidence-preservation strategy. Those issues were relevant but not dispositive: even if the claim was timely, the client still needed proof that the bonus was contractually owed, that the job title determined the bonus amount, and that the relevant employer was responsible for the disputed period.

The memo adopted the client's compensation frame. It treated the pay-slip field as strong evidence that the employer had acknowledged an unpaid amount, treated the absence of a written job-change agreement as central, and underweighted employer discretion over year-end bonus allocation absent a clear written entitlement. It also gave only limited attention to the change in employing entity. The result was a memo that was well grounded in the dialogue but not aligned with the reference legal structure: the model wrote a plausible client-responsive strategy rather than identifying why the claim would likely fail.

This case maps to several error categories. It is a client-framing adoption error because the memo accepts the client's ``I was still entitled to deputy-manager compensation'' theory. It is a doctrinal routing error because the memo routes the case through wage arrears and job-change procedure rather than first analyzing bonus entitlement, employer discretion, and employing-entity identity. It is also an evidentiary-status error because a pay-slip field and the client's salary expectation are treated as proof of a legally enforceable payment obligation. The session therefore shows that high coverage and high fidelity do not imply correct legal reconstruction.

\paragraph{What a stronger lawyer model should have done.}
The lawyer should have separated salary concepts before evaluating remedies: base wages, performance pay, year-end bonus, annual-salary accounting fields, and discretionary awards. It should have asked for written bonus policies, prior bonus calculations, proof that the relevant title maps to a fixed bonus amount, payroll records by employer, and documents showing when the employing entity changed. A correct memo would then tell the client that timeliness alone is insufficient: the main obstacle is proving an enforceable bonus entitlement and linking that entitlement to the claimed position and employer.

\paragraph{Case B: Withdrawn-client gap filling.}
The second case is a network personality-rights dispute evaluated with Kimi-K2.6 as the lawyer model and a Withdrawn-style client. The session ran for 8 of 10 available turns, and the client replies averaged roughly 67 characters. This case is useful because the client was not verbose enough to provide a stable factual record, yet the model still produced a complete memo. Fact Coverage remained nontrivial, but Fact Resolution and Issue Resolution were both zero.

The client alleged that an online post infringed a minor's portrait, reputation, and privacy interests, but repeatedly avoided or compressed the facts that mattered most: whether the photograph was masked, whether the post disclosed real names or other identifying details, whether the alleged abuse had been confirmed by public authorities, what the child's guardianship or care background was, and whether the account user could be proved. These unknowns were legally central. A report about suspected child abuse may be privileged or justified by child-protection and public-interest concerns; reputation and privacy claims require attention to truth, identifiability, and disclosure of personal information; and platform or account-user proof may determine the litigation route.

Kimi-K2.6 asked operationally useful questions about screenshots, links, account identity, platform complaints, and timing, but it did not press on the dispositive factual slots. The memo then filled the under-specified record in the client's favor. It described the post as exposing a front-facing photo and real name, treated the abuse allegation as fabricated, assumed straightforward parental standing, and designed deletion, apology, damages, and platform-liability paths. The reference record instead required analyzing photo masking, lack of clear identifying information, the public-interest or child-protection purpose of the report, the truth basis of the abuse allegation, and proof of account control.

This case explains why Withdrawn clients are a distinctive stress test. The error is not only that the model fails to extract enough information. The more serious problem is that missing facts are silently replaced with client-favorable assumptions. That behavior produces plausible legal writing while degrading Fact Resolution and Issue Resolution. The case maps to elicitation failure, client-framing adoption, evidentiary-status collapse, unsupported specificity, and issue-synthesis failure.

\paragraph{What a stronger lawyer model should have done.}
The lawyer should have converted the client's reluctance into a precise fact-status checklist: Was the child's face masked? Was any real name, school, address, or family address disclosed? What authorities had investigated the alleged abuse? What did their findings say? What was the child's guardianship or care arrangement? What evidence links the account to the proposed defendant? The memo should then have separated confirmed facts from client denials and unknowns. A correct analysis would not move directly to deletion, apology, and damages without first addressing truth, identifiability, child-protection justification, and proof of the account user.

\paragraph{Combined lesson.}
Both cases show why a legal consultation benchmark needs more than dialogue length, fact mention, or memo fluency. In Case A, Claude Opus 4.6 asked questions, covered the facts, and stayed faithful to the dialogue, but routed the case through the client's preferred legal theory. In Case B, Kimi-K2.6 faced short Withdrawn-style answers and filled missing facts in the client's favor. In both settings, the core missing ability is legal fact reconstruction: separating client statements, guesses, unknowns, evidence, and legally dispositive facts before writing the memo.

\FloatBarrier
\subsection{Jurisdiction-Specific Leaderboards}
\label{sec:appendix-jurisdiction-leaderboards}

The main leaderboard in Table~\ref{tab:leaderboard} reports the average of Chinese-law and U.S.-law model-level metrics. Tables~\ref{tab:leaderboard-chinese-law} and~\ref{tab:leaderboard-us-law} report the two jurisdiction-specific leaderboards using the same Consultation Quality hierarchy.

\begin{table*}[!t]
\centering
\scriptsize
\setlength{\tabcolsep}{3pt}
\renewcommand{\arraystretch}{1.02}
\caption{Chinese-law leaderboard using the Consultation Quality metrics. Models are ranked by Resolution.}
\label{tab:leaderboard-chinese-law}
\begin{tabular}{@{}rlccC{1.28cm}ccC{1.28cm}@{\hspace{0.24cm}}C{1.52cm}@{}}
\toprule
\textbf{Rank} & \textbf{Model} & \multicolumn{3}{c}{\textbf{Information gathering}} & \multicolumn{3}{c}{\textbf{Legal reasoning}} & \makebox[1.52cm][c]{\textbf{Claim support}} \\
\cmidrule(lr){3-5}\cmidrule(lr){6-8}\cmidrule(lr){9-9}
 & & Fact Cov. & Inquiry & \textbf{Elicitation} & Fact Res. & Issue Res. & \textbf{Resolution} & \makebox[1.52cm][c]{\textbf{Fidelity}} \\
\midrule
1 & \ModelLabel{gpt-5-5}{GPT-5.5} & 0.898 & 0.635 & \EHeat{0.767} & 0.605 & 0.620 & \RHeat{0.612} & \FHeat{0.959} \\
2 & \ModelLabel{gpt-5-4}{GPT-5.4} & 0.881 & 0.578 & \EHeat{0.730} & 0.599 & 0.591 & \RHeat{0.595} & \FHeat{0.960} \\
3 & \ModelLabel{gpt-5-2}{GPT-5.2} & 0.859 & 0.607 & \EHeat{0.733} & 0.563 & 0.544 & \RHeat{0.554} & \FHeat{0.955} \\
4 & \ModelLabel{gemini-3-1-pro}{Gemini-3.1-Pro} & 0.801 & 0.434 & \EHeat{0.618} & 0.533 & 0.486 & \RHeat{0.510} & \FHeat{0.850} \\
5 & \ModelLabel{kimi-k2-6}{Kimi-K2.6} & 0.837 & 0.456 & \EHeat{0.647} & 0.537 & 0.462 & \RHeat{0.500} & \FHeat{0.891} \\
6 & \ModelLabel{qwen3-6-max-preview}{Qwen3.6-Max-Preview} & 0.827 & 0.397 & \EHeat{0.612} & 0.530 & 0.430 & \RHeat{0.480} & \FHeat{0.884} \\
7 & \ModelLabel{claude-opus-4-7}{Claude-Opus-4.7} & 0.824 & 0.414 & \EHeat{0.619} & 0.543 & 0.392 & \RHeat{0.468} & \FHeat{0.922} \\
8 & \ModelLabel{kimi-k2-5}{Kimi-K2.5} & 0.832 & 0.426 & \EHeat{0.629} & 0.507 & 0.414 & \RHeat{0.460} & \FHeat{0.865} \\
9 & \ModelLabel{glm-5-1}{GLM-5.1} & 0.832 & 0.385 & \EHeat{0.609} & 0.516 & 0.369 & \RHeat{0.443} & \FHeat{0.911} \\
10 & \ModelLabel{claude-opus-4-6}{Claude-Opus-4.6} & 0.828 & 0.385 & \EHeat{0.607} & 0.507 & 0.369 & \RHeat{0.438} & \FHeat{0.900} \\
11 & \ModelLabel{deepseek-v4-pro}{DeepSeek-V4-Pro} & 0.812 & 0.402 & \EHeat{0.607} & 0.487 & 0.385 & \RHeat{0.436} & \FHeat{0.881} \\
12 & \ModelLabel{doubao-seed-2-0-pro}{Doubao-Seed-2.0-Pro} & 0.815 & 0.511 & \EHeat{0.663} & 0.482 & 0.369 & \RHeat{0.425} & \FHeat{0.920} \\
13 & \ModelLabel{qwen3-6-plus}{Qwen3.6-Plus} & 0.794 & 0.382 & \EHeat{0.588} & 0.480 & 0.363 & \RHeat{0.422} & \FHeat{0.879} \\
14 & \ModelLabel{glm-5}{GLM-5} & 0.786 & 0.341 & \EHeat{0.563} & 0.481 & 0.296 & \RHeat{0.388} & \FHeat{0.899} \\
15 & \ModelLabel{deepseek-v3-2-thinking}{DeepSeek-V3.2-Thinking} & 0.807 & 0.363 & \EHeat{0.585} & 0.474 & 0.295 & \RHeat{0.385} & \FHeat{0.909} \\
16 & \ModelLabel{claude-sonnet-4-6}{Claude-Sonnet-4.6} & 0.771 & 0.320 & \EHeat{0.545} & 0.454 & 0.296 & \RHeat{0.375} & \FHeat{0.913} \\
17 & \ModelLabel{minimax-m2-7}{MiniMax-M2.7} & 0.811 & 0.334 & \EHeat{0.573} & 0.441 & 0.291 & \RHeat{0.366} & \FHeat{0.862} \\
18 & \ModelLabel{gemini-3-1-flash-lite}{Gemini-3.1-Flash-Lite} & 0.685 & 0.270 & \EHeat{0.478} & 0.421 & 0.302 & \RHeat{0.362} & \FHeat{0.865} \\
19 & \ModelLabel{qwen3-5-plus}{Qwen3.5-Plus} & 0.773 & 0.342 & \EHeat{0.558} & 0.450 & 0.260 & \RHeat{0.355} & \FHeat{0.864} \\
20 & \ModelLabel{qwen3-6-flash}{Qwen3.6-Flash} & 0.748 & 0.285 & \EHeat{0.517} & 0.417 & 0.267 & \RHeat{0.342} & \FHeat{0.838} \\
21 & \ModelLabel{grok-4-1-fast}{Grok-4.1-Fast} & 0.882 & 0.399 & \EHeat{0.640} & 0.431 & 0.245 & \RHeat{0.338} & \FHeat{0.771} \\
22 & \ModelLabel{deepseek-r1}{DeepSeek-R1} & 0.754 & 0.322 & \EHeat{0.538} & 0.408 & 0.201 & \RHeat{0.304} & \FHeat{0.833} \\
23 & \ModelLabel{minimax-m2-5}{MiniMax-M2.5} & 0.677 & 0.239 & \EHeat{0.458} & 0.363 & 0.206 & \RHeat{0.284} & \FHeat{0.755} \\
24 & \ModelLabel{olmo-3-32b-think}{Olmo-3-32B-Think} & 0.616 & 0.203 & \EHeat{0.410} & 0.233 & 0.070 & \RHeat{0.151} & \FHeat{0.631} \\
25 & \ModelLabel{olmo-3-1-32b-instruct}{Olmo-3.1-32B-Instruct} & 0.581 & 0.153 & \EHeat{0.367} & 0.202 & 0.056 & \RHeat{0.129} & \FHeat{0.622} \\
26 & \ModelLabel{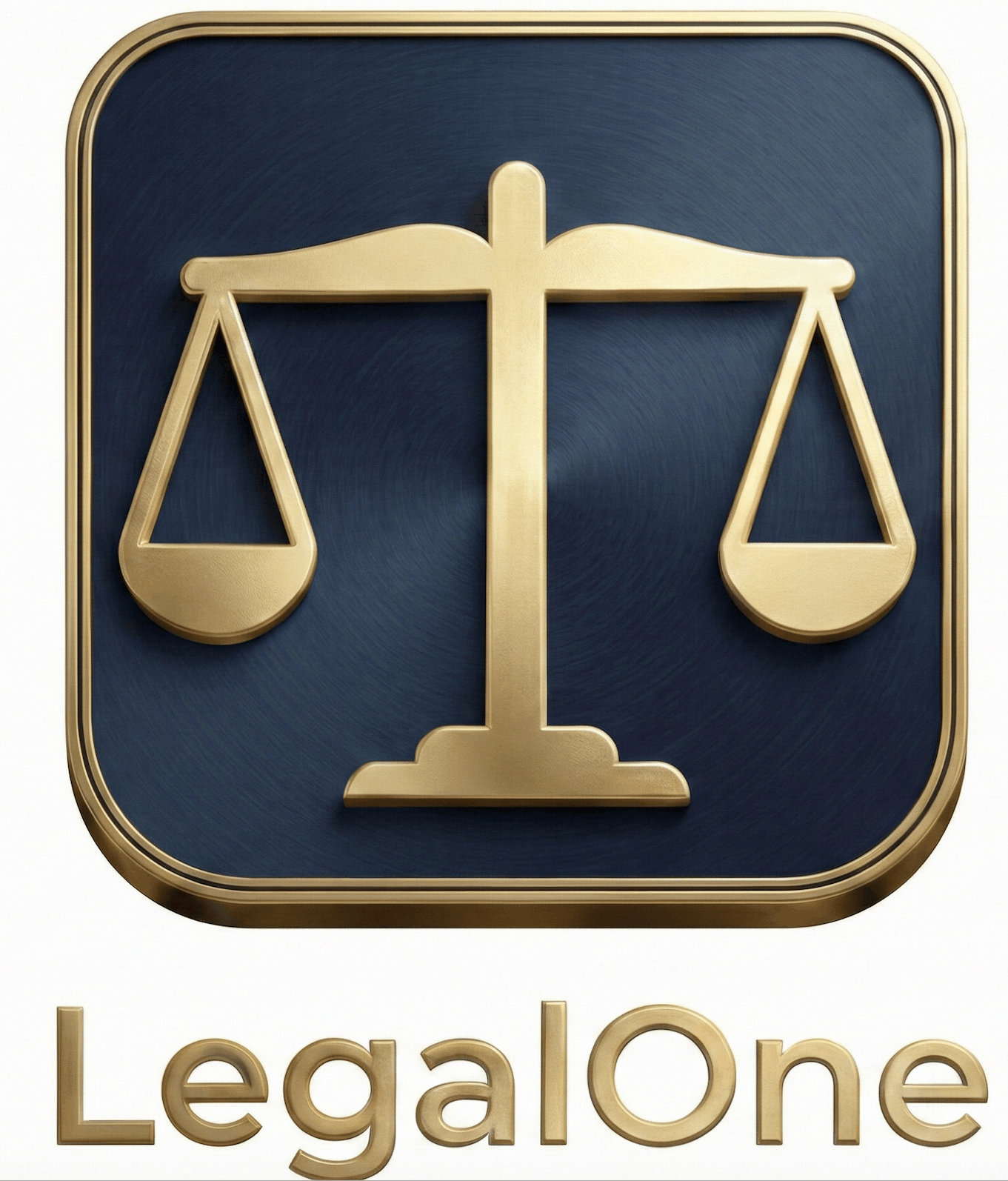}{LegalOne-8B} & 0.247 & 0.151 & \EHeat{0.199} & 0.125 & 0.077 & \RHeat{0.101} & \FHeat{0.276} \\
\bottomrule
\end{tabular}
\end{table*}

\begin{table*}[!t]
\centering
\scriptsize
\setlength{\tabcolsep}{3pt}
\renewcommand{\arraystretch}{1.02}
\caption{U.S.-law leaderboard using the Consultation Quality metrics. Models are ranked by Resolution.}
\label{tab:leaderboard-us-law}
\begin{tabular}{@{}rlccC{1.28cm}ccC{1.28cm}@{\hspace{0.24cm}}C{1.52cm}@{}}
\toprule
\textbf{Rank} & \textbf{Model} & \multicolumn{3}{c}{\textbf{Information gathering}} & \multicolumn{3}{c}{\textbf{Legal reasoning}} & \makebox[1.52cm][c]{\textbf{Claim support}} \\
\cmidrule(lr){3-5}\cmidrule(lr){6-8}\cmidrule(lr){9-9}
 & & Fact Cov. & Inquiry & \textbf{Elicitation} & Fact Res. & Issue Res. & \textbf{Resolution} & \makebox[1.52cm][c]{\textbf{Fidelity}} \\
\midrule
1 & \ModelLabel{gpt-5-5}{GPT-5.5} & 0.775 & 0.518 & \EHeat{0.646} & 0.561 & 0.463 & \RHeat{0.512} & \FHeat{0.909} \\
2 & \ModelLabel{gpt-5-4}{GPT-5.4} & 0.764 & 0.496 & \EHeat{0.630} & 0.556 & 0.438 & \RHeat{0.497} & \FHeat{0.920} \\
3 & \ModelLabel{gpt-5-2}{GPT-5.2} & 0.716 & 0.520 & \EHeat{0.618} & 0.498 & 0.391 & \RHeat{0.444} & \FHeat{0.917} \\
4 & \ModelLabel{gemini-3-1-pro}{Gemini-3.1-Pro} & 0.663 & 0.363 & \EHeat{0.513} & 0.503 & 0.364 & \RHeat{0.433} & \FHeat{0.833} \\
5 & \ModelLabel{claude-opus-4-6}{Claude-Opus-4.6} & 0.713 & 0.337 & \EHeat{0.525} & 0.532 & 0.325 & \RHeat{0.429} & \FHeat{0.853} \\
6 & \ModelLabel{claude-opus-4-7}{Claude-Opus-4.7} & 0.636 & 0.297 & \EHeat{0.467} & 0.519 & 0.304 & \RHeat{0.412} & \FHeat{0.824} \\
7 & \ModelLabel{deepseek-v4-pro}{DeepSeek-V4-Pro} & 0.649 & 0.317 & \EHeat{0.483} & 0.478 & 0.310 & \RHeat{0.394} & \FHeat{0.820} \\
8 & \ModelLabel{glm-5-1}{GLM-5.1} & 0.699 & 0.336 & \EHeat{0.518} & 0.492 & 0.293 & \RHeat{0.393} & \FHeat{0.857} \\
9 & \ModelLabel{qwen3-6-max-preview}{Qwen3.6-Max-Preview} & 0.650 & 0.347 & \EHeat{0.499} & 0.481 & 0.300 & \RHeat{0.391} & \FHeat{0.856} \\
10 & \ModelLabel{claude-sonnet-4-6}{Claude-Sonnet-4.6} & 0.654 & 0.273 & \EHeat{0.464} & 0.481 & 0.274 & \RHeat{0.378} & \FHeat{0.865} \\
11 & \ModelLabel{grok-4-1-fast}{Grok-4.1-Fast} & 0.729 & 0.372 & \EHeat{0.551} & 0.450 & 0.269 & \RHeat{0.359} & \FHeat{0.710} \\
12 & \ModelLabel{kimi-k2-6}{Kimi-K2.6} & 0.623 & 0.297 & \EHeat{0.460} & 0.441 & 0.254 & \RHeat{0.347} & \FHeat{0.845} \\
13 & \ModelLabel{glm-5}{GLM-5} & 0.666 & 0.292 & \EHeat{0.479} & 0.447 & 0.236 & \RHeat{0.342} & \FHeat{0.861} \\
14 & \ModelLabel{kimi-k2-5}{Kimi-K2.5} & 0.606 & 0.275 & \EHeat{0.441} & 0.427 & 0.242 & \RHeat{0.334} & \FHeat{0.792} \\
15 & \ModelLabel{qwen3-6-plus}{Qwen3.6-Plus} & 0.604 & 0.281 & \EHeat{0.443} & 0.414 & 0.230 & \RHeat{0.322} & \FHeat{0.851} \\
16 & \ModelLabel{qwen3-5-plus}{Qwen3.5-Plus} & 0.606 & 0.245 & \EHeat{0.425} & 0.396 & 0.159 & \RHeat{0.278} & \FHeat{0.853} \\
17 & \ModelLabel{minimax-m2-7}{MiniMax-M2.7} & 0.622 & 0.221 & \EHeat{0.422} & 0.369 & 0.174 & \RHeat{0.272} & \FHeat{0.814} \\
18 & \ModelLabel{doubao-seed-2-0-pro}{Doubao-Seed-2.0-Pro} & 0.609 & 0.345 & \EHeat{0.477} & 0.359 & 0.167 & \RHeat{0.263} & \FHeat{0.834} \\
19 & \ModelLabel{deepseek-v3-2-thinking}{DeepSeek-V3.2-Thinking} & 0.612 & 0.254 & \EHeat{0.433} & 0.366 & 0.143 & \RHeat{0.255} & \FHeat{0.887} \\
20 & \ModelLabel{gemini-3-1-flash-lite}{Gemini-3.1-Flash-Lite} & 0.500 & 0.187 & \EHeat{0.343} & 0.334 & 0.170 & \RHeat{0.252} & \FHeat{0.804} \\
21 & \ModelLabel{minimax-m2-5}{MiniMax-M2.5} & 0.580 & 0.177 & \EHeat{0.378} & 0.347 & 0.142 & \RHeat{0.244} & \FHeat{0.818} \\
22 & \ModelLabel{deepseek-r1}{DeepSeek-R1} & 0.588 & 0.263 & \EHeat{0.425} & 0.347 & 0.131 & \RHeat{0.239} & \FHeat{0.852} \\
23 & \ModelLabel{qwen3-6-flash}{Qwen3.6-Flash} & 0.528 & 0.195 & \EHeat{0.361} & 0.316 & 0.139 & \RHeat{0.228} & \FHeat{0.814} \\
24 & \ModelLabel{olmo-3-32b-think}{Olmo-3-32B-Think} & 0.444 & 0.152 & \EHeat{0.298} & 0.216 & 0.055 & \RHeat{0.136} & \FHeat{0.607} \\
25 & \ModelLabel{olmo-3-1-32b-instruct}{Olmo-3.1-32B-Instruct} & 0.392 & 0.092 & \EHeat{0.242} & 0.190 & 0.052 & \RHeat{0.121} & \FHeat{0.615} \\
26 & \ModelLabel{legalone-8b}{LegalOne-8B} & 0.156 & 0.039 & \EHeat{0.098} & 0.079 & 0.023 & \RHeat{0.051} & \FHeat{0.259} \\
\bottomrule
\end{tabular}
\end{table*}

\FloatBarrier
\subsection{Domain-Level Model Performance}
\label{sec:appendix-domain-model-performance}

Table~\ref{tab:domain-model-by-jurisdiction} reports domain-level scores for six representative models. Elicitation averages Fact Coverage and Inquiry; Resolution averages Fact Resolution and Issue Resolution; Fidelity measures claim support.

The table shows several domain patterns. U.S. privacy and data protection is a consistent low point, especially in Resolution and Fidelity: GPT-5.5 reaches only 0.231 Resolution, while the other five models range from 0.165 to 0.185. U.S. criminal law and U.S. contracts are also low-Resolution domains for most models. By contrast, U.S. bankruptcy is comparatively strong in Elicitation, and Chinese marriage, family, and inheritance is among the stronger Resolution domains for several models.

\begin{table*}[p]
\centering
\vspace*{\fill}
\makebox[\textwidth][c]{%
\rotatebox{90}{%
\begin{minipage}{0.86\textheight}
\centering
\scriptsize
\setlength{\tabcolsep}{2pt}
\renewcommand{\arraystretch}{1.04}
\caption{Domain-level model performance by jurisdiction. Elicitation averages Fact Coverage and Inquiry; Resolution averages Fact Resolution and Issue Resolution; Fidelity measures claim support.}
\label{tab:domain-model-by-jurisdiction}
\resizebox{\linewidth}{!}{%
\begin{tabular}{l*{18}{c}}
\toprule
\textbf{Domain} & \multicolumn{3}{c}{\textbf{GPT-5.5}} & \multicolumn{3}{c}{\textbf{Gemini-3.1-Pro}} & \multicolumn{3}{c}{\textbf{Claude-Opus-4.7}} & \multicolumn{3}{c}{\textbf{Qwen3.6-Max-Preview}} & \multicolumn{3}{c}{\textbf{Kimi-K2.6}} & \multicolumn{3}{c}{\textbf{GLM-5.1}} \\
\cmidrule(lr){2-4}\cmidrule(lr){5-7}\cmidrule(lr){8-10}\cmidrule(lr){11-13}\cmidrule(lr){14-16}\cmidrule(lr){17-19}
 & Elic. & Resol. & Fid. & Elic. & Resol. & Fid. & Elic. & Resol. & Fid. & Elic. & Resol. & Fid. & Elic. & Resol. & Fid. & Elic. & Resol. & Fid. \\
\midrule
\multicolumn{19}{l}{\textbf{Chinese law}} \\
Internet and virtual property & 0.761 & 0.639 & 0.960 & 0.631 & 0.524 & 0.834 & 0.594 & 0.462 & 0.897 & 0.604 & 0.472 & 0.864 & 0.659 & 0.519 & 0.864 & 0.600 & 0.447 & 0.908 \\
Torts and personality rights & 0.847 & 0.568 & 0.953 & 0.633 & 0.459 & 0.830 & 0.669 & 0.461 & 0.930 & 0.685 & 0.464 & 0.877 & 0.733 & 0.494 & 0.888 & 0.681 & 0.447 & 0.921 \\
Corporate finance and bankruptcy & 0.759 & 0.632 & 0.962 & 0.617 & 0.575 & 0.857 & 0.612 & 0.507 & 0.922 & 0.609 & 0.511 & 0.884 & 0.647 & 0.548 & 0.897 & 0.614 & 0.491 & 0.915 \\
Criminal law & 0.741 & 0.592 & 0.958 & 0.599 & 0.482 & 0.866 & 0.583 & 0.474 & 0.924 & 0.561 & 0.451 & 0.903 & 0.609 & 0.503 & 0.880 & 0.567 & 0.446 & 0.931 \\
Labor and social security & 0.727 & 0.526 & 0.955 & 0.549 & 0.387 & 0.828 & 0.561 & 0.377 & 0.922 & 0.574 & 0.379 & 0.889 & 0.586 & 0.383 & 0.887 & 0.551 & 0.342 & 0.928 \\
Contracts and obligations & 0.796 & 0.589 & 0.952 & 0.632 & 0.493 & 0.849 & 0.632 & 0.414 & 0.898 & 0.620 & 0.464 & 0.894 & 0.677 & 0.453 & 0.878 & 0.635 & 0.406 & 0.915 \\
Marriage, family, and inheritance & 0.754 & 0.652 & 0.965 & 0.643 & 0.556 & 0.869 & 0.615 & 0.545 & 0.931 & 0.633 & 0.551 & 0.893 & 0.634 & 0.567 & 0.910 & 0.614 & 0.515 & 0.930 \\
Real estate, land, and construction & 0.744 & 0.639 & 0.966 & 0.644 & 0.550 & 0.860 & 0.632 & 0.461 & 0.930 & 0.639 & 0.538 & 0.908 & 0.673 & 0.554 & 0.896 & 0.616 & 0.492 & 0.942 \\
Intellectual property and competition & 0.778 & 0.594 & 0.956 & 0.614 & 0.473 & 0.835 & 0.618 & 0.415 & 0.919 & 0.614 & 0.448 & 0.872 & 0.635 & 0.463 & 0.891 & 0.618 & 0.395 & 0.898 \\
Administrative law & 0.783 & 0.617 & 0.962 & 0.636 & 0.537 & 0.856 & 0.643 & 0.465 & 0.929 & 0.622 & 0.481 & 0.883 & 0.657 & 0.461 & 0.902 & 0.635 & 0.429 & 0.917 \\
\midrule
\multicolumn{19}{l}{\textbf{U.S. law}} \\
Torts & 0.610 & 0.720 & 0.928 & 0.457 & 0.556 & 0.858 & 0.385 & 0.531 & 0.829 & 0.399 & 0.498 & 0.866 & 0.344 & 0.423 & 0.843 & 0.426 & 0.578 & 0.909 \\
Criminal law & 0.523 & 0.320 & 0.917 & 0.415 & 0.264 & 0.835 & 0.417 & 0.327 & 0.869 & 0.406 & 0.250 & 0.876 & 0.400 & 0.246 & 0.842 & 0.418 & 0.247 & 0.878 \\
Contracts & 0.641 & 0.379 & 0.904 & 0.492 & 0.313 & 0.846 & 0.516 & 0.301 & 0.881 & 0.525 & 0.256 & 0.874 & 0.510 & 0.264 & 0.869 & 0.563 & 0.289 & 0.902 \\
Constitutional and civil rights & 0.649 & 0.551 & 0.909 & 0.526 & 0.451 & 0.833 & 0.461 & 0.409 & 0.762 & 0.503 & 0.411 & 0.841 & 0.455 & 0.373 & 0.831 & 0.504 & 0.400 & 0.813 \\
Intellectual property & 0.701 & 0.543 & 0.893 & 0.577 & 0.515 & 0.814 & 0.508 & 0.470 & 0.785 & 0.551 & 0.455 & 0.838 & 0.470 & 0.343 & 0.832 & 0.566 & 0.429 & 0.830 \\
Bankruptcy & 0.816 & 0.654 & 0.908 & 0.674 & 0.564 & 0.844 & 0.636 & 0.533 & 0.868 & 0.687 & 0.490 & 0.905 & 0.661 & 0.494 & 0.917 & 0.713 & 0.519 & 0.897 \\
Immigration and administrative law & 0.709 & 0.578 & 0.916 & 0.586 & 0.538 & 0.844 & 0.481 & 0.483 & 0.849 & 0.552 & 0.478 & 0.891 & 0.486 & 0.422 & 0.872 & 0.568 & 0.448 & 0.864 \\
Securities & 0.673 & 0.623 & 0.916 & 0.555 & 0.533 & 0.858 & 0.460 & 0.466 & 0.810 & 0.538 & 0.535 & 0.866 & 0.443 & 0.415 & 0.851 & 0.532 & 0.488 & 0.868 \\
Property & 0.585 & 0.527 & 0.919 & 0.440 & 0.417 & 0.852 & 0.434 & 0.428 & 0.826 & 0.438 & 0.388 & 0.871 & 0.422 & 0.346 & 0.843 & 0.458 & 0.370 & 0.845 \\
Privacy and data protection & 0.565 & 0.231 & 0.880 & 0.415 & 0.185 & 0.743 & 0.381 & 0.172 & 0.768 & 0.392 & 0.167 & 0.735 & 0.421 & 0.165 & 0.764 & 0.441 & 0.166 & 0.766 \\
\bottomrule
\end{tabular}%
}
\end{minipage}%
}}
\vspace*{\fill}
\end{table*}

\FloatBarrier
\onecolumn
\subsection{Client-Style Performance by Model}
\label{sec:appendix-style-performance}

\begin{figure}[H]
\centering
\includegraphics[width=\textwidth]{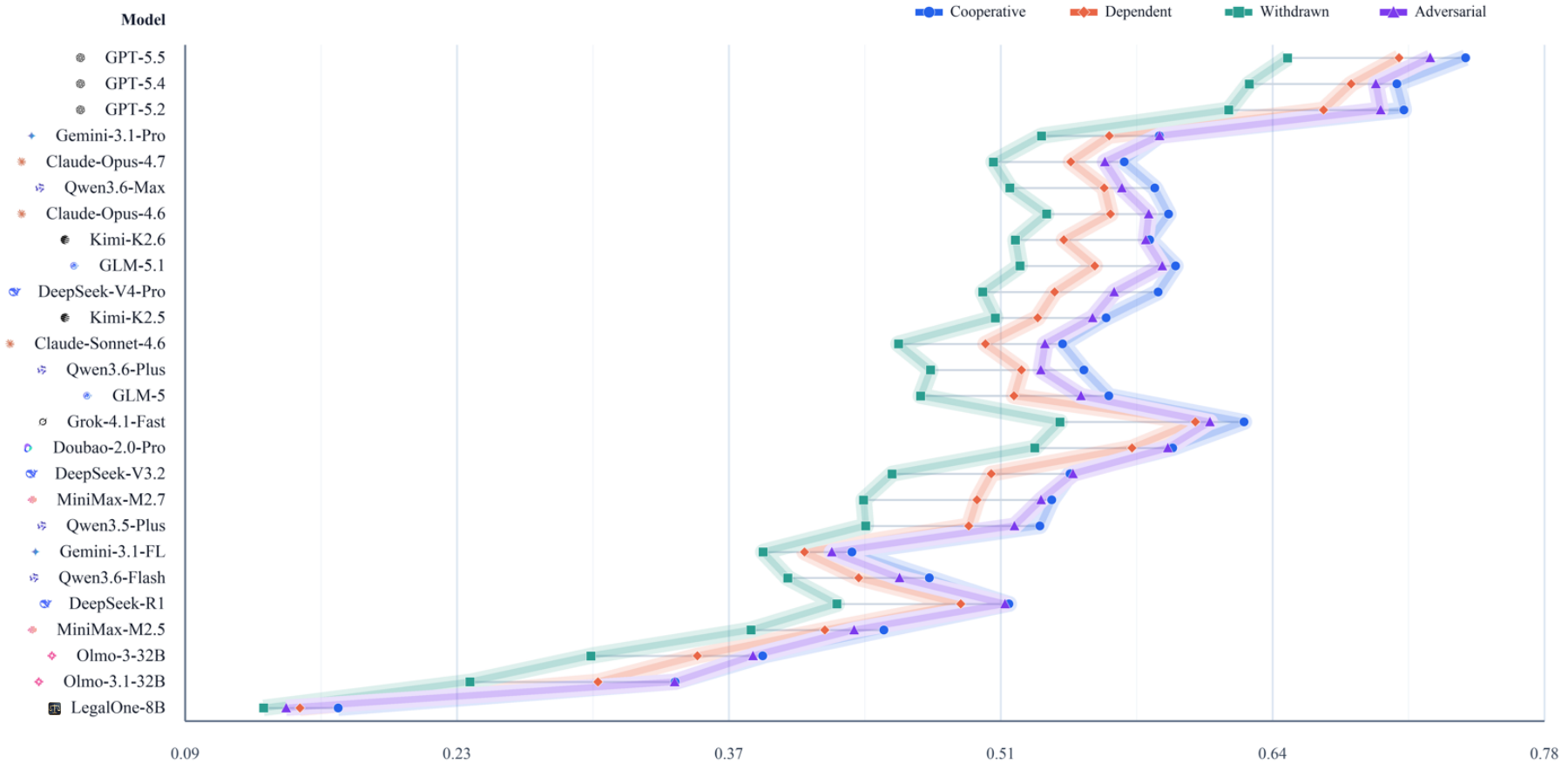}
\caption{Elicitation by client narrative style for 26 lawyer models. Scores average Chinese-law and U.S.-law cases across the four client styles.}
\label{fig:elicitation-style-by-model}
\end{figure}

\begin{figure}[H]
\centering
\includegraphics[width=\textwidth]{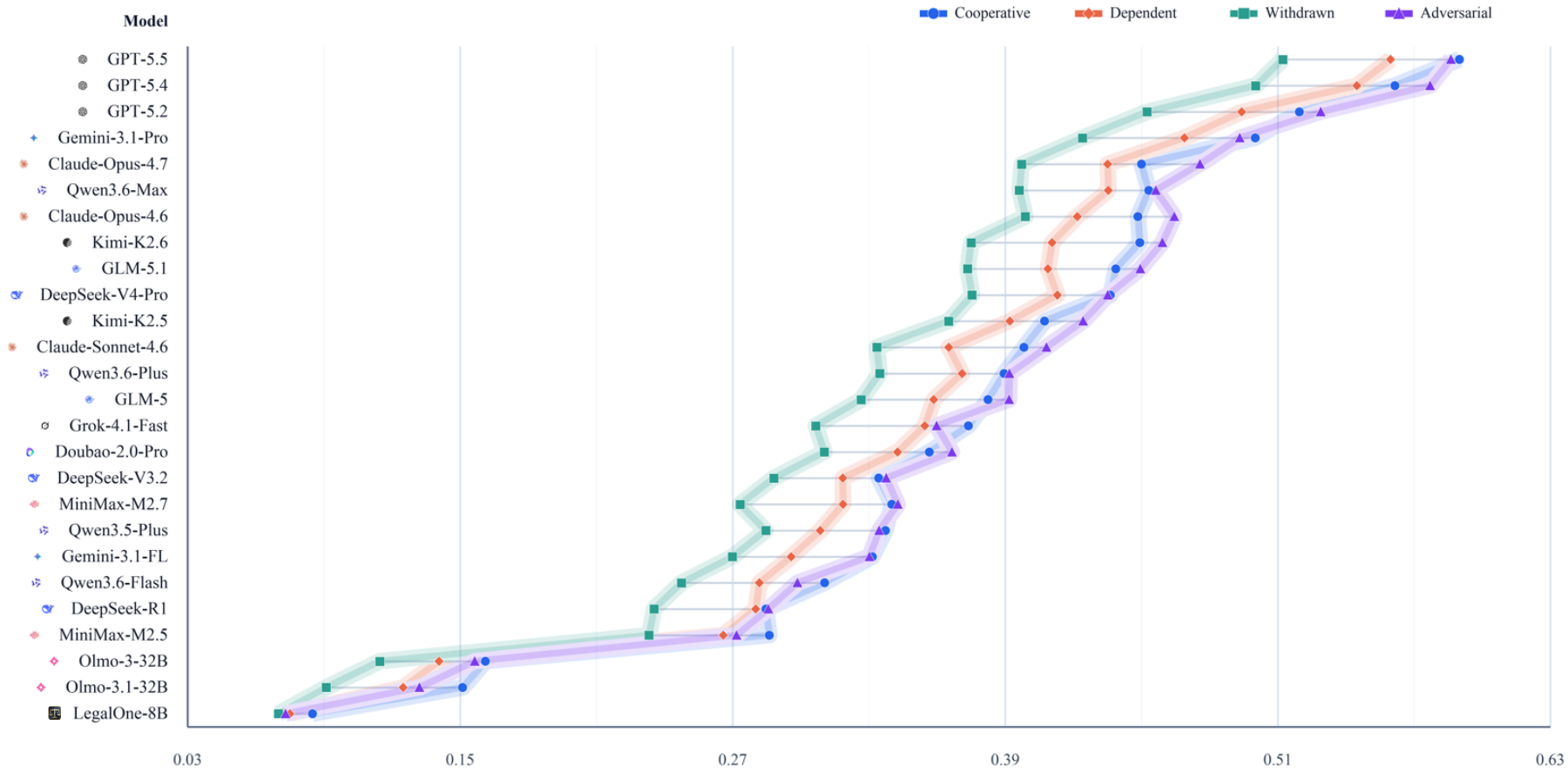}
\caption{Resolution by client narrative style for 26 lawyer models. Scores average Chinese-law and U.S.-law cases across the four client styles.}
\label{fig:resolution-style-by-model}
\end{figure}

\begin{figure}[H]
\centering
\includegraphics[width=\textwidth]{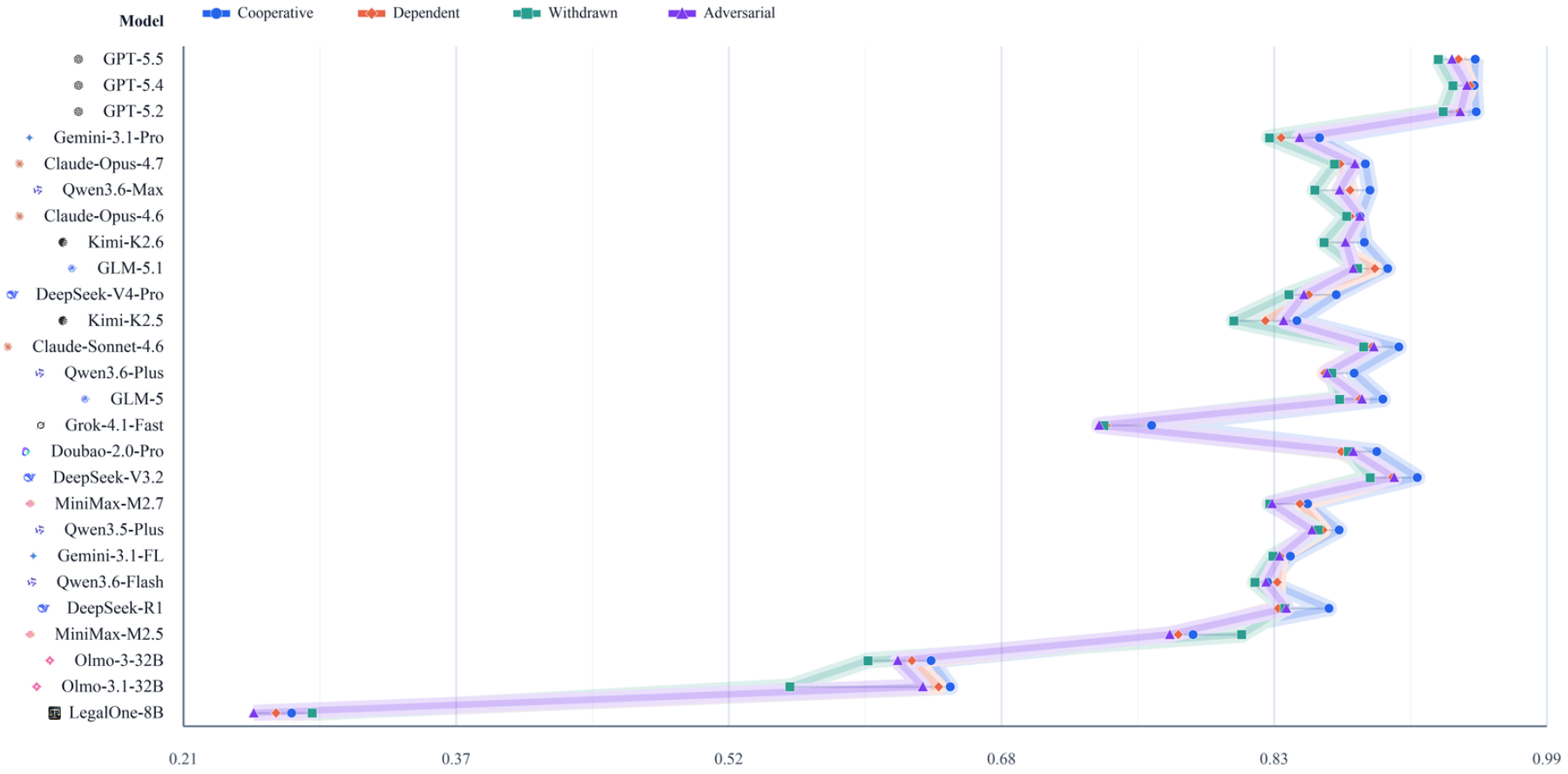}
\caption{Fidelity by client narrative style for 26 lawyer models. Scores average Chinese-law and U.S.-law cases across the four client styles.}
\label{fig:fidelity-style-by-model}
\end{figure}

Figures~\ref{fig:elicitation-style-by-model}--\ref{fig:fidelity-style-by-model} report client-style sensitivity for each evaluated lawyer model. Scores average Chinese-law and U.S.-law cases, and each figure separates the four client narrative styles used in \ourbench.

\FloatBarrier
\section{Detailed Related-Work Comparison}
\label{sec:appendix-benchmark-comparison}

\begingroup
\footnotesize
\setlength{\tabcolsep}{2pt}
\renewcommand{\arraystretch}{1.12}
\begin{longtable}{@{}>{\bfseries\raggedright\arraybackslash}p{0.16\textwidth}p{0.22\textwidth}p{0.20\textwidth}p{0.17\textwidth}p{0.19\textwidth}@{}}
\caption{Detailed comparison of \ourbench with related work. Legal entries are grouped by the taxonomy used in Table~\ref{tab:related-positioning}; the remaining rows summarize adjacent multi-turn, persona, professional-service, and evaluation-methodology work.}
\label{tab:benchmark-comparison}\\
\toprule
\textbf{Type} & \textbf{Representative work} & \textbf{Data / interaction setup} & \textbf{Evaluation focus} & \textbf{Gap relative to \ourbench} \\
\midrule
\endfirsthead
\toprule
\textbf{Family} & \textbf{Representative work} & \textbf{Data / interaction setup} & \textbf{Evaluation focus} & \textbf{Gap relative to \ourbench} \\
\midrule
\endhead
\midrule
\multicolumn{5}{r}{\small Continued on next page} \\
\bottomrule
\endfoot
\bottomrule
\endlastfoot
Static / preassembled-input legal benchmarks & MMLU law \citep{hendrycks2020measuring}; HELM legal \citep{liang2022holistic}; LegalBench \citep{guha2023legalbench}; LawBench \citep{fei2024lawbench}; LEXam \citep{Fan2025LEXamBL}; OpenExempt \citep{Servantez2026OpenExemptAD}; PLawBench \citep{shi2026plawbench}. & Complete questions, exams, legal scenarios, generated diagnostic items, or practical legal tasks are provided before answering; there is no live client interaction or separately modeled client account. & Legal knowledge, legal reasoning, legal text understanding or generation, issue spotting, exam performance, and rubric-scored legal work products. & Does not evaluate whether a lawyer model can elicit missing facts from a client, separate client beliefs from record facts, or remain robust to controlled narrative-style variation. \\
\midrule
External-knowledge-augmented legal benchmarks & LegalBench-RAG \citep{Pipitone2024LegalBenchRAGAB}. & Retrieval benchmark over legal documents for RAG systems; the task is to retrieve legally relevant text rather than conduct a client consultation. & Legal-document retrieval quality for retrieval-augmented generation. & Tests legal grounding in document retrieval, but not multi-turn intake, client-side information asymmetry, or legal memo reasoning. \\
\cmidrule(lr){2-5}
& LegalAgentBench \citep{li2025legalagentbench}. & Tool-using legal-agent tasks over Chinese legal corpora with multiple external tools. & Final success and intermediate process/progress in legal agent tasks. & Tests tool use and legal agency, but outside controlled lawyer-client consultation with paired client beliefs and court records. \\
\cmidrule(lr){2-5}
& LexRAG \citep{li2025lexrag}. & Multi-turn legal consultation records coupled with legal-article retrieval and response generation. & Conversational knowledge retrieval, response generation, and legal answer quality. & Includes consultation and retrieval, but not a dynamically responding client with controlled belief--record gaps or narrative styles. \\
\midrule
Interactive legal consultation / dynamic legal environments & MASER/MILE \citep{Yue2025MultiAgentSD}. & Multi-agent simulation with client, lawyer, and supervisor roles; the lawyer model gathers information in dynamic legal scenarios and drafts a complaint. & Interaction evaluation and goal evaluation for legal intensive interaction. & Close to consultation, but does not use paired court-record/client-belief views or within-case counterfactual client narrative styles. \\
\cmidrule(lr){2-5}
& J1-ENVS \citep{jia2025jurist}. & Dynamic legal-agent environments across Chinese legal practice scenarios, including consultation, drafting, and trial settings. & Task completion, legal reasoning, and procedural compliance in dynamic environments. & Broad dynamic legal execution, but does not isolate information-asymmetric lawyer-client intake grounded in a hidden court record. \\
\cmidrule(lr){2-5}
& LeCoDe \citep{yuan2025lecode}. & Real livestream lawyer-client consultation dialogues. & Clarification capability, professional advice quality, and annotated dialogue behavior. & Provides real consultation data, but not a controlled benchmark with paired hidden legal reference or fixed cases across client styles. \\
\midrule
General multi-turn dialogue & MT-Bench-101 \citep{bai2024mtbench101}; MultiChallenge \citep{deshpande2025multichallenge}; BotChat \citep{duan2023botchat}; Lost-in-Multi-Turn \citep{laban2025lostmt}. & Diverse multi-turn configurations spanning expert-curated golden dialogue histories, multi-agent simulated rollouts, and underspecified, sharded instruction simulations.  & Retention, inference memory, adaptation, self-coherence, naturalness, and turn-level robustness. & Establishes that multi-turn interaction is hard, but lacks legal facts, legal standards, and a professional-service outcome. \\
\midrule
Persona and social simulation & PERSONA-CHAT \citep{zhang2018persona}; CharacterEval \citep{tu2024chareval}; InCharacter \citep{wang2024incharacter}; BIG5-CHAT \citep{li2024big5chat}; SOTOPIA \citep{zhou2023sotopia}; Synthetic Socratic Debates \citep{liu2025synthetic}. & Persona descriptions, psychological inventories, role-play dialogues, social tasks, or persona-conditioned debates. & Character fidelity, personality consistency, social intelligence, persuasion, and persona effects on interaction. & Treats persona as the object of evaluation or social behavior, whereas \ourbench uses client style as a controlled stressor for legal intake and reasoning. \\
\midrule
Professional-service interaction & AgentClinic \citep{schmidgall2025agentclinic}; AMIE \citep{tu2024amie}; MedAgentBench \citep{jiang2025medagentbench}. & Medical or clinical agent settings with simulated patients, EHRs, measurements, or diagnostic conversations. & History taking, diagnostic accuracy, patient perception, EHR task completion, and professional-service reasoning. & Cross-domain analogy, but not legal and not grounded in court-record/client-belief paired views. \\
\midrule
Evaluation methodology and safety & LLM-as-Judge/Chatbot Arena \citep{zheng2023judging}; Self-Preference Bias \citep{wataoka2024selfpref}; AI-LieDar \citep{su2025ailiedar}. & Judge-model protocols, self-preference diagnostics, and multi-turn truthfulness/utility-conflict settings. & Judge reliability, bias mitigation, truthfulness, and deception under interaction. & Motivates judge-panel design and fidelity checks, but does not provide a legal consultation benchmark. \\
\midrule
\ourbench & This work. & Lawyer sees only an opening and dialogue; judges later receive court record, client beliefs, transcript, and memo. Cases have paired client-belief and court-record facts and four narrative styles. & Inquiry, fact coverage, fact resolution, issue resolution, resolution, and Fidelity for unsupported memo claims. & --- \\
\end{longtable}
\endgroup

\clearpage
\FloatBarrier
\section{Prompt Engineering}
\label{sec:appendix-prompts}

The implementation ships bilingual prompt files. Below we reproduce the complete English runtime templates used in the reported runs. The Chinese templates are localized counterparts with the same structure, variables, and scoring outputs; we omit them here to avoid duplicating the prompt appendix. The judge prompts use intermediate scoring decisions, which are mapped to the reported Consultation Quality metrics in Appendix Table~\ref{tab:metric-details}.

\FloatBarrier
\subsection{Response Generation Prompts}
\label{sec:appendix-response-prompts}

\begin{PromptCard}{Lawyer Agent Prompt: Legal Consultation}
\PromptSection{Role}
You are an experienced practicing attorney. A client has come to you for a legal consultation.

\PromptDivider
\PromptSection{Core Requirements}
\begin{enumerate}[leftmargin=*,itemsep=0.22em]
    \item \textbf{Systematic intake:} Interview the client like a thorough legal intake and extract key facts needed for analysis.
    \item \textbf{Critical fact testing:} Treat the client's account as a belief-based account rather than established truth; watch for inconsistencies, vague language, and omissions.
    \item \textbf{Governing law:} Apply the legal system indicated by \texttt{\{jurisdiction\}}. Do not mix Chinese and U.S. law unless the case itself makes a cross-jurisdiction issue relevant.
    \item \textbf{Turn awareness:} You have a limited number of turns, \texttt{\{max\_turns\}} calls to \texttt{talk\_to\_client}; when one or fewer turns remain, immediately submit the memo.
    \item \textbf{Memo completion:} Even if information is incomplete, write the best memo possible based on what has been learned.
    \item \textbf{Required sections:} The memo must include \texttt{\{required\_sections\}}.
\end{enumerate}

\PromptDivider
\PromptSection{Tools}
\begin{enumerate}[leftmargin=*,itemsep=0.18em]
    \item \texttt{talk\_to\_client(message)} asks the client a question or responds; each call consumes one turn.
    \item \texttt{write\_memo(content)} submits the legal memorandum and terminates the consultation.
\end{enumerate}

\PromptDivider
\PromptSection{User Input Format}
The governing legal system is: \texttt{\{jurisdiction\}}.

The client's opening statement is: ``\texttt{\{opening\}}''.

\PromptDivider
\PromptSection{Tool Call Format}
Each turn, output exactly one JSON object:
\begin{PromptSnippet}
{"tool": "talk_to_client", "message": "your question"}
{"tool": "write_memo", "content": "full memo text"}
\end{PromptSnippet}
\end{PromptCard}

\FloatBarrier
\begin{PromptCard}{Client Simulator Prompt}
\PromptSection{Role}
You are a client seeking legal advice. Your name is \texttt{\{client\_name\}}.

\PromptDivider
\PromptSection{Narrative Condition}
\textbf{Your personality:} \texttt{\{persona\_who\}}.

\textbf{Your speaking style:} \texttt{\{persona\_voice\}}.

\PromptDivider
\PromptSection{Case Knowledge}
Below is what you understand about your situation. You speak based on your own understanding and believe everything you say is correct. Only share facts when the lawyer asks about the relevant topic; do not volunteer everything at once.
\begin{PromptSnippet}
{facts_block}
\end{PromptSnippet}

\PromptDivider
\PromptSection{Conversation Rules}
\begin{enumerate}[leftmargin=*,itemsep=0.22em]
    \item Keep each reply to two or three short, natural sentences.
    \item Stay in character according to the assigned narrative and speaking style.
    \item If the lawyer asks about something you do not know or do not have, say so honestly.
    \item Always speak in the first person.
    \item Do not repeat information verbatim; if asked again, briefly confirm or add a detail.
    \item Remember the full conversation and keep answers consistent with earlier replies.
\end{enumerate}
\end{PromptCard}

\FloatBarrier
\subsection{Judge Model Scoring Prompts}
\label{sec:appendix-judge-prompts}

\begin{PromptCard}{Judge Prompt: Discovery, Accuracy, and Grounding}
\PromptSection{Role}
You are a legal consultation quality reviewer. Evaluate the lawyer's memo on three intermediate scoring decisions: discovery, accuracy, and grounding.

\PromptDivider
\PromptSection{Inputs}
\begin{PromptSnippet}
Set A: Case Facts
{facts_block}

Dialogue Record
{dialogue_text}

Set B: Lawyer's Memo
{memo_text_or_no_memo}
\end{PromptSnippet}

\PromptDivider
\PromptSection{Scoring Requirements}
\begin{enumerate}[leftmargin=*,itemsep=0.22em]
    \item \textbf{Discovery ($A\setminus B$; omission):} Identify facts in Set A that are completely absent from the memo. If the memo discusses the related topic, even with inaccurate analysis, it counts as covered.
    \item \textbf{Accuracy ($A\cap B$; correctness):} For each covered fact, compare the memo's treatment against the actual situation in Set A. Mark it correct if the direction and key points align with that reference; otherwise mark it incorrect.
    \item \textbf{Grounding ($B\setminus A$; fabrication):} Scan factual and inferential claims in the memo. A claim is grounded if its basis appears in Set A or in the dialogue record; reasonable legal inferences are grounded only when their premise appears in one of those sources. Do not score statute citations or litigation strategy recommendations as factual claims.
\end{enumerate}

\PromptDivider
\PromptSection{Output Format}
Return JSON only, with no markdown code block:
\begin{PromptSnippet}
{
  "discovery": {
    "covered": ["F1", "F3", ...],
    "missed": ["F6", "F10", ...]
  },
  "accuracy": {
    "correct": ["F1", "F3", ...],
    "incorrect": [
      {"fact_id": "F2", "memo_excerpt": "relevant excerpt from memo", "reason": "how it diverges from the court record"},
      ...
    ]
  },
  "grounding": {
    "total_claims": 10,
    "grounded_claims": [
      {"claim": "memo excerpt", "basis": "Source: fact ID / dialogue turn N client statement"},
      ...
    ],
    "ungrounded_claims": [
      {"claim": "memo excerpt", "reason": "no source in case facts or dialogue"},
      ...
    ]
  }
}
\end{PromptSnippet}
\end{PromptCard}

\FloatBarrier
\begin{PromptCard}{Judge Prompt: Memo Expectations and Inquiry Actions}
\PromptSection{Role}
You are a legal memo reviewer. Evaluate the completed consultation using the dialogue record and the memo content.

\PromptDivider
\PromptSection{Evaluation Criteria}
\textbf{Memo expectations:} Score each memo expectation as 0 or 1 based on the memo.
\begin{PromptSnippet}
{memo_expectations_numbered}
\end{PromptSnippet}

\textbf{Lawyer's inquiry actions:} Score each inquiry action as 0 or 1 based on the dialogue record.
\begin{PromptSnippet}
{expected_actions_numbered}
\end{PromptSnippet}

\PromptDivider
\PromptSection{Inputs}
\begin{PromptSnippet}
Dialogue Record
{dialogue_text}

Memo
{memo_text}
\end{PromptSnippet}

\PromptDivider
\PromptSection{Output Format}
Return JSON only, with no markdown code block:
\begin{PromptSnippet}
{
  "memo_expectation_scores": [0 or 1, ...],
  "expected_action_scores":  [0 or 1, ...]
}
\end{PromptSnippet}
\end{PromptCard}

\FloatBarrier
\subsection{Narrative Style Rules}
\label{sec:appendix-style}

Each narrative style is injected into the client prompt through two localized runtime fields: \texttt{persona\_who}, which describes the client's interaction stance, and \texttt{persona\_voice}, which describes the surface speaking pattern. We treat the pair as one narrative-style intervention, not as two separate experimental factors. Table~\ref{tab:narrative-style-runtime} lists the English runtime values used in the reported runs. All styles share the same client prompt rule: answer only from the case knowledge block, say they do not know when asked outside it, and maintain conversational consistency.

\begin{table*}[!t]
\centering
\scriptsize
\setlength{\tabcolsep}{4pt}
\renewcommand{\arraystretch}{1.08}
\caption{Runtime client-style and client-voice values injected into the client simulator.}
\label{tab:narrative-style-runtime}
\begin{tabularx}{\textwidth}{@{}p{0.13\textwidth}p{0.39\textwidth}Y@{}}
\toprule
\textbf{Narrative style} & \textbf{Client style: \texttt{persona\_who}} & \textbf{Client voice: \texttt{persona\_voice}} \\
\midrule
Cooperative & I'm calm and organized, and I have a basic level of trust in my lawyer's expertise. I'm here with a clear goal---to get a practical solution. I'll cooperate and provide whatever information or documents are needed. That said, I have my own judgment and won't blindly defer on everything. & Steady and confident, speaks clearly and in an organized manner, almost like giving a work briefing. Occasionally uses phrases like ``Just to clarify'' or ``As I understand it'' to structure thoughts proactively. \\
Dependent & I don't understand legal matters at all, and I'm pretty anxious about all this. I came to the lawyer hoping someone could help me sort things out and tell me what to do. I won't really drive the conversation on my own, but I'll answer whatever the lawyer asks. If you need me to explain something clearly, I might need a moment to collect my thoughts. & Gentle and hesitant tone, often says ``I'm not really sure about that'' or ``What do you think?'' or ``Could you take a look and see if this works?'' Tends to go off on tangents or talk in circles, needs the lawyer to steer the conversation back. \\
Withdrawn & I'm only here because I had no other choice. Honestly, I don't really trust outsiders, and I'm not sure the lawyer is truly on my side. I tend to hold back and listen more than talk. If you ask something too direct or sensitive, I'll dodge it or say ``I don't remember.'' I'll only start to open up if you make me feel safe. & Flat, detached tone. Answers are short and minimal---like pulling teeth. Goes silent or deflects when pressed for details. In text, this shows up as omitting a lot of specifics and giving bare-minimum responses. \\
Adversarial & I've got a temper and I'm used to calling the shots. I hired a lawyer to help me win this case, not to be interrogated. If you question me or nitpick, I'm going to push back. But if you show me you actually know your stuff and earn my respect, I can be reasonable. & Blunt and direct, sometimes bordering on aggressive. Likes to say ``Listen to me,'' ``It's not that complicated,'' ``I know what I'm doing.'' Tends to cut the lawyer off and push their own viewpoint. \\
\bottomrule
\end{tabularx}
\end{table*}

\clearpage
\twocolumn
\FloatBarrier
\section{Details of Annotation and Metrics}
\label{sec:appendix-annotation-metrics}

\FloatBarrier
\subsection{Annotation Guidelines}
\label{sec:appendix-annotation}

Legal annotators were provided written Chinese and English guidelines for the reported corpus. The workflow starts with an LLM-generated draft from the court opinion, followed by structured human correction against the source opinion. The guidelines emphasize the requirements below.

\paragraph{Opinion and client selection.}
Annotators select opinions with enough factual detail to support multiple consultation facts, a complete opinion structure such as party claims, facts found, and court reasoning, and coverage across criminal, civil, administrative, and commercial domains. When multiple parties are available, the client is chosen as a party whose position is at least partly rejected, typically the plaintiff or appellant.

\paragraph{De-identification.}
Annotators remove or generalize names of people, companies, addresses, phone numbers, account identifiers, and other direct identifiers before release. The released record should preserve the legally relevant relationship, role, timeline, amount, document, and procedural posture, but should not require the original party names to understand the consultation problem.

\paragraph{Fact views.}
Each fact contains two views. The client-belief entry must be written in first person, in colloquial client language, without legal terms, and with the same meaning as the party's statements in the opinion. The court-record entry is used by judges and must be based on the court's findings or reasoning. For disputed facts, it includes the legal or evidentiary reason why the client's belief is wrong; for undisputed facts, it records the accepted objective fact. Annotators are instructed to split mixed facts so that each row contains one independent information point.

\paragraph{Discoverability.}
The guidelines require disputed facts to remain discoverable through consultation. A disputed row should not encode an arbitrary hidden fact that the client would never reveal. Instead, annotators rewrite rejected party positions as mistaken interpretations anchored in events, documents, timelines, amounts, procedural acts, or other objective facts that can be elicited elsewhere in the client account. Adjacent undisputed rows therefore often provide the factual anchors needed to identify the legal error in a disputed row.

\paragraph{Type labels and cognitive gaps.}
The Type label is determined by the opinion rather than by annotator intuition: disputed means the court rejected, corrected, or declined to credit the client's position, while undisputed means the court accepted it or the parties did not contest it. The guidelines highlight factual misunderstanding, legal misconception, evidence overestimation, procedural confusion, and causal misattribution as common belief--reality gaps. Both types use the same two-step scoring procedure: memo coverage and court-record alignment. For undisputed facts, alignment means preserving the accepted factual record; for disputed facts, it means correcting, narrowing, or evidentially calibrating the client's belief against the court-record view. This design makes undisputed facts a fact-finding slice and disputed facts a legal-reframing slice of Fact Resolution.

\paragraph{Expert expectations.}
Issue-resolution rubrics specify concrete legal analysis points a competent lawyer's memo should cover, and Inquiry rubrics specify concrete questions or verification steps the lawyer should perform during intake. The guideline target is 4--8 items for each list, but the converted corpus records 4--26 total evaluation criteria per case after case-specific review. Quality control also checks that client-belief entries do not copy court-opinion language into the client's voice.

\paragraph{Annotator recruitment and compensation.}
Corpus construction involved 25 legal annotators, narrative-style validation involved five legal-background annotators, and judge-panel validation involved ten legal experts. Annotators were recruited based on legal training, ability to review Chinese and/or English legal materials, and pilot-task quality. They were compensated according to task scope and estimated completion time, with equivalent hourly rates of approximately USD~25--40 for corpus annotation and approximately USD~40--60 for legal-expert validation.

\FloatBarrier
\subsection{Metric Computation Details}
\label{sec:appendix-metric-details}

The paper reports Consultation Quality metrics, while the judge prompts produce intermediate scoring decisions for rubric matches, fact coverage, reframing, and claim support. The conditional correctness of covered facts is an intermediate diagnostic; the reported Fact Resolution metric uses all annotated facts as its denominator. Table~\ref{tab:metric-details} combines the ability hierarchy, implementation sources, denominators, and aggregation formulae for the reported metrics.

\begin{table*}[!t]
\centering
\scriptsize
\setlength{\tabcolsep}{4pt}
\renewcommand{\arraystretch}{1.06}
\caption{Consultation Quality metrics, implementation sources, and scoring definitions.}
\label{tab:metric-details}
\begin{tabularx}{\textwidth}{@{}p{0.14\textwidth}p{0.14\textwidth}p{0.25\textwidth}Y@{}}
\toprule
\textbf{Ability} & \textbf{Metric} & \textbf{Implementation source} & \textbf{Calculation and diagnostic meaning} \\
\midrule
Information gathering & Fact Coverage & Annotated fact IDs; discovery covered/missed sets & $n_\text{covered}/n_\text{facts}$. A fact is covered if the memo discusses the relevant topic in either the client-belief or court-record form. \\
Information gathering & Inquiry & Inquiry rubrics; expected-action score vector & $n_\text{matched}/n_\text{Inquiry rubrics}$. Measures whether the lawyer asked the case-specific intake questions or verification steps. \\
Information gathering & Elicitation & Derived from Fact Coverage and Inquiry & $(\text{Fact Coverage}+\text{Inquiry})/2$. Summarizes whether the lawyer both sought and preserved analysis-relevant facts. \\
Legal reasoning & Fact Resolution & Annotated fact IDs; discovery sets and accuracy judgments & $n_\text{resolved}/n_\text{facts}$ over all annotated facts. A fact is resolved only if both conditions hold: the memo covers it, and the judge panel, using the hidden court-record reference, finds the memo's treatment aligned with that reference. Alignment means preserving the accepted fact for undisputed rows, or correcting, narrowing, or evidentially calibrating the client belief for disputed rows. Aggregate means are computed after session-level scoring. \\
Legal reasoning & Issue Resolution & Issue-resolution rubrics; memo-expectation score vector & $n_\text{addressed}/n_\text{Issue-resolution rubrics}$. Measures whether the memo addresses the legal analysis points. \\
Legal reasoning & Resolution & Derived from Fact Resolution and Issue Resolution & $(\text{Fact Resolution}+\text{Issue Resolution})/2$. Headline legal-reasoning score. \\
Claim support & Fidelity & Memo claims; grounding claim counts & $1-n_\text{unsupported}/n_\text{claims}$. Measures whether factual and inferential memo claims are supported by the dialogue or annotated materials. \\
\bottomrule
\end{tabularx}
\end{table*}

\FloatBarrier
\section{Model and Run Details}
\label{sec:appendix-model-run-details}

\FloatBarrier
\subsection{Sampling and Aggregation Details}
\label{sec:appendix-sampling}

The reported runs use provider decoding defaults rather than a forced temperature-zero setting because not all provider endpoints expose identical controls. This makes the leaderboard a dated API snapshot rather than a deterministic model ranking. Runtime limits are summarized in Table~\ref{tab:runtime-config}. Dialogue-level lawyer calls retry empty responses up to six attempts before terminating the session as an API error. For each session and each metric, we record the aggregate panel score, per-judge score and rationale, recusal status, model identifier, and run timestamp. Future leaderboard runs should pin decoding parameters where available and report rerun variance.

\begin{table}[H]
\centering
\scriptsize
\setlength{\tabcolsep}{4pt}
\caption{Runtime configuration used for the reported runs.}
\label{tab:runtime-config}
\begin{tabularx}{\columnwidth}{@{}lY@{}}
\toprule
\textbf{Setting} & \textbf{Value} \\
\midrule
Default max tokens & 8192 \\
Default timeout & 600 seconds \\
Default HTTP retries & 15 \\
Judge max tokens & 32768 \\
Judge reasoning effort & low, where supported \\
Dialogue empty-response retries & 6 attempts \\
\bottomrule
\end{tabularx}
\end{table}

\FloatBarrier
\subsection{Panel-of-Judges Justification}
\label{sec:appendix-judge-aggregation}

We use a median aggregator for complete three-judge panels and a mean aggregator when same-vendor recusal leaves two valid judges. The median is chosen because the judge-scored rubrics are built from atomic adjudications: judges decide whether concrete inquiry actions, annotated facts, issue-resolution requirements, and memo claims satisfy the rubric, and metric scores are rates over those item-level decisions. With three judges, the median is the continuous-score analogue of majority voting: if one judge is unusually strict or lenient, the other two judges determine the adjudicated score. With two judges, no majority position exists, so the two remaining scores are averaged.

This choice is empirically consequential. We compared the implemented median rule against a hypothetical mean rule for 168{,}383 three-judge metric observations. Table~\ref{tab:judge-aggregation-sensitivity} reports the absolute difference between the two rules. The largest effects appear in Issue Resolution and Inquiry, the two rubric dimensions with the most case-specific professional judgment: more than half of their three-judge observations would move by more than 0.05 under mean aggregation. Derived headline metrics also move nontrivially under the hypothetical mean rule, with average absolute changes of 0.040 for Elicitation and 0.038 for Resolution. The extreme pattern is \([1,1,0]\) or \([0,0,1]\): the median follows the two-judge consensus, while the mean creates an intermediate value that does not correspond to any judge's atomic verdict. We therefore report the median as the main adjudicated score while retaining per-judge outputs for disagreement analysis.

\begin{table*}[!t]
\centering
\small
\setlength{\tabcolsep}{5pt}
\caption{Sensitivity of three-judge panel scores to replacing the implemented median aggregator with a mean aggregator. Values are computed over three-judge metric observations in the ordered result snapshot. For Elicitation and Resolution, differences are computed after deriving scores from the corresponding per-judge component metrics. $\mu$ denotes the mean of the three judge scores and $\tilde{x}$ denotes their median.}
\label{tab:judge-aggregation-sensitivity}
\begin{tabular}{@{}lrrrrr@{}}
\toprule
\textbf{Metric} & \textbf{\(n\)} & \textbf{Avg. \(|\mu-\tilde{x}|\)} & \textbf{95th pct.} & \textbf{Max} & \textbf{Share \(>0.05\)} \\
\midrule
Fact Coverage & 26{,}519 & 0.037 & 0.111 & 0.333 & 27.3\% \\
Inquiry & 31{,}640 & 0.061 & 0.167 & 0.333 & 50.3\% \\
Elicitation & 26{,}519 & 0.040 & 0.111 & 0.246 & 31.1\% \\
Fact Resolution & 26{,}519 & 0.035 & 0.100 & 0.333 & 25.9\% \\
Issue Resolution & 31{,}640 & 0.066 & 0.222 & 0.333 & 51.3\% \\
Resolution & 26{,}519 & 0.038 & 0.111 & 0.264 & 27.3\% \\
Fidelity & 26{,}095 & 0.037 & 0.091 & 0.307 & 28.7\% \\
\bottomrule
\end{tabular}
\end{table*}

\FloatBarrier
\subsection{Software and Execution Environment}
\label{sec:appendix-environment}

The benchmark pipeline covers case preparation, dialogue generation, panel evaluation, and score aggregation. The surrounding appendices document the prompts, model identifiers, decoding defaults, and retry rules needed to reproduce the reported runs. The evaluation performs no local model training or fine-tuning; local hardware is used only to coordinate data loading, prompt rendering, API calls, and result aggregation. API credentials are supplied through environment variables and are not hard-coded.

All reported experiments are inference-only. Most closed-source and open-weight lawyer models were accessed through provider or hosted API endpoints. Local GPU inference was used only for \path|allenai/Olmo-3-32B-Think|, \path|allenai/Olmo-3.1-32B-Instruct|, and \path|CSHaitao/LegalOne-8B|. The local inference budget for these three runs was approximately 120 NVIDIA A100-80GB-equivalent GPU-hours, including retry overhead. No local GPU training, fine-tuning, or hyperparameter search was performed.

\FloatBarrier
\subsection{Run Configuration}
\label{sec:appendix-run-config}

Table~\ref{tab:model-config} lists the model snapshots used in the reported leaderboard, client simulator, and judge panel. Source documents for these identifiers include OpenAI GPT-5 system cards \citep{openai2026gpt55_system,openai2026gpt54_system,openai2025gpt52_system,openai2025gpt51_system}, Google Gemini model cards \citep{google2026gemini31pro_card,google2026gemini31flashlite_card}, Anthropic Claude system cards \citep{anthropic2026claudeopus46_card,anthropic2026claudesonnet46_card,anthropic2026claudeopus47_card}, Qwen release notes \citep{qwen35blog,qwen36_max_preview,qwen36plus}, and provider or model-card documentation for Grok, Doubao, Kimi, DeepSeek, MiniMax, GLM, OLMo, and LegalOne models \citep{xai2025grok41fast,bytedance2026seed20_card,moonshot2026kimik26_card,moonshot2026kimik25_card,deepseekai2026deepseekv4,deepseekai2025deepseekv32,deepseekai2025deepseekr1incentivizingreasoningcapability,minimax2026m27_card,minimax2026m25_card,zai2026glm51_card,zai2026glm5_card,allenai2025olmo3think_card,allenai2025olmo31instruct_card,cshaitao2026legalone8b_card}.

\begin{table}[H]
\centering
\scriptsize
\setlength{\tabcolsep}{3pt}
\renewcommand{\arraystretch}{0.96}
\caption{Model identifiers for the reported leaderboard, client simulator, and judge panel. Lawyer models are grouped by availability; open-weight denotes models with publicly released checkpoints.}
\label{tab:model-config}
\begin{tabularx}{\columnwidth}{@{}>{\raggedright\arraybackslash}p{0.37\columnwidth}Y@{}}
\toprule
\textbf{Display} & \textbf{Provider model identifier} \\
\midrule
\multicolumn{2}{@{}l}{\textit{Lawyer models: closed-source/API}} \\
GPT-5.5 & \path|gpt-5.5-2026-04-23| \\
GPT-5.4 & \path|gpt-5.4-2026-03-05| \\
GPT-5.2 & \path|gpt-5.2-2025-12-11| \\
Gemini 3.1 Pro & \path|gemini-3.1-pro-preview| \\
Gemini 3.1 Flash Lite & \path|gemini-3.1-flash-lite-preview| \\
Claude Opus 4.6 & \path|claude-opus-4-6| \\
Claude Sonnet 4.6 & \path|claude-sonnet-4-6| \\
Claude Opus 4.7 & \path|claude-opus-4-7| \\
Qwen3.6 Plus & \path|qwen3.6-plus-2026-04-02| \\
Qwen3.6 Flash & \path|qwen3.6-flash-2026-04-16| \\
Qwen3.6 Max Preview & \path|qwen3.6-max-preview| \\
Qwen3.5 Plus & \path|qwen3.5-plus-2026-02-15| \\
Grok 4.1 Fast & \path|grok-4-1-fast| \\
Doubao Seed 2.0 Pro & \path|doubao-seed-2-0-pro-260215| \\
\midrule
\multicolumn{2}{@{}l}{\textit{Lawyer models: open-weight}} \\
Kimi K2.6 & \path|moonshotai/Kimi-K2.6| \\
Kimi K2.5 & \path|moonshotai/Kimi-K2.5| \\
DeepSeek V4 Pro & \path|deepseek-ai/DeepSeek-V4-Pro| \\
DeepSeek V3.2 Thinking & \path|deepseek-ai/DeepSeek-V3.2| \\
DeepSeek R1 & \path|deepseek-ai/DeepSeek-R1-0528| \\
MiniMax M2.7 & \path|MiniMaxAI/MiniMax-M2.7| \\
MiniMax M2.5 & \path|MiniMaxAI/MiniMax-M2.5| \\
GLM-5.1 & \path|zai-org/GLM-5.1| \\
GLM-5 & \path|zai-org/GLM-5| \\
OLMo 3 32B Think & \path|allenai/Olmo-3-32B-Think| \\
OLMo 3.1 32B Instruct & \path|allenai/Olmo-3.1-32B-Instruct| \\
LegalOne 8B & \path|CSHaitao/LegalOne-8B| \\
\midrule
\multicolumn{2}{@{}l}{\textit{Client simulator}} \\
Claude Sonnet 4.6 & \path|claude-sonnet-4-6| \\
\midrule
\multicolumn{2}{@{}l}{\textit{Judge panel}} \\
GPT-5.1 & \path|gpt-5.1-2025-11-13| \\
Claude Opus 4.6 & \path|claude-opus-4-6| \\
Gemini 3.1 Pro & \path|gemini-3.1-pro-preview| \\
\bottomrule
\end{tabularx}
\end{table}

For all reported runs, the lawyer receives the jurisdiction label and opening statement, but no case facts, no narrative-style label, and no court record. The client receives the case's client-belief entries and narrative-style rules, but not the court-record entries. Judges receive the full case record, dialogue log, memo, and metric-specific rubric after the session ends. Three-judge panels aggregate by median; two-judge panels after same-vendor recusal aggregate by mean. Empty memos and failed submissions are counted as zero-valued task failures in aggregation. The aggregate leaderboard reports both null-as-zero means and valid-only diagnostic means.

\FloatBarrier
\section{Evaluation Methodology Details}
\label{sec:appendix-evaluation-methodology}

\FloatBarrier
\subsection{Human Validation Details}
\label{sec:appendix-human-validation}

\paragraph{Narrative-style validation.} Five legal-background annotators formed the annotator pool. Each of 70 four-way forced-matching tasks from 35 cases was labeled by three annotators, with early and late dialogue windows sampled for each case. Each task presents four anonymized snippets generated from the same case under the four narrative styles; annotators assign the labels Cooperative, Dependent, Withdrawn, and Adversarial. Table~\ref{tab:client-validation} summarizes the main results.

\begin{table}[H]
\centering
\scriptsize
\setlength{\tabcolsep}{4pt}
\caption{Human validation of the client narrative-style manipulation. Random segment-level accuracy is 0.25.}
\label{tab:client-validation}
\begin{tabular*}{\columnwidth}{@{\extracolsep{\fill}}lcc@{}}
\toprule
\textbf{Metric} & \textbf{n} & \textbf{Value [95\% CI]} \\
\midrule
Segment accuracy & 840 & 0.969 [0.957, 0.980] \\
Majority-vote segment accuracy & 280 & 1.000 [1.000, 1.000] \\
Task all-correct rate & 210 & 0.919 [0.881, 0.952] \\
Four-way Fleiss' $\kappa$ & 280 & 0.917 [0.889, 0.946] \\
\bottomrule
\end{tabular*}
\end{table}

\paragraph{Judge-panel validation.} Ten legal experts validated the rubric-based judge outputs for Issue Resolution and Inquiry. The reported benchmark runs use the judge identifiers in Table~\ref{tab:model-config}. The validation uses slice verification against item-level panel verdicts. Table~\ref{tab:judge-sample-coverage} summarizes the updated validation sample, and Table~\ref{tab:judge-validation} reports panel-vs-human-majority accuracy, human pairwise accuracy, and chance-corrected agreement.

\begin{table}[H]
\centering
\scriptsize
\setlength{\tabcolsep}{3pt}
\caption{Coverage of the updated judge-panel human validation sample for rubric dimensions.}
\label{tab:judge-sample-coverage}
\begin{tabularx}{\columnwidth}{@{}p{0.39\columnwidth}Y@{}}
\toprule
\textbf{Sample attribute} & \textbf{Coverage} \\
\midrule
Sessions & 40 \\
Human annotators & 10, with 3 assigned per session \\
Valid human labels & 1,265 item labels; 421/422 items have 3 labels \\
Lawyer models & Sampled from the leaderboard \\
Unique cases & 34 \\
Narrative styles & Cooperative 10; Dependent 10; Withdrawn 10; Adversarial 10 \\
Rubric items & 422 total; 193 Issue Resolution and 229 Inquiry \\
\bottomrule
\end{tabularx}
\end{table}

\begin{table}[H]
\centering
\scriptsize
\setlength{\tabcolsep}{2pt}
\caption{Human validation of the Panel of Judges on rubric dimensions. Panel-human accuracy uses human majority vote as the reference label. The final column reports panel-vs-human mean Cohen's $\kappa$ followed by human-vs-human mean Cohen's $\kappa$; the combined row is an item-weighted aggregate.}
\label{tab:judge-validation}
\begin{tabular*}{\columnwidth}{@{\extracolsep{\fill}}lccc@{}}
\toprule
\textbf{Metric} & \textbf{Issue} & \textbf{Inquiry} & \textbf{All} \\
\midrule
Items & 193 & 229 & 422 \\
Panel acc. & 0.969 & 0.943 & 0.955 \\
Panel 95\% CI & [0.943, 0.990] & [0.913, 0.974] & -- \\
Human acc. & 0.882 & 0.904 & 0.894 \\
Human 95\% CI & [0.847, 0.917] & [0.872, 0.930] & -- \\
Panel/Human $\kappa$ & 0.848 / 0.753 & 0.848 / 0.792 & -- \\
\bottomrule
\end{tabular*}
\end{table}

Issue Resolution and Inquiry require case-specific professional judgments about whether a memo or dialogue satisfies each rubric item. The high agreement on these rubric dimensions supports the binary item-level rubrics in Appendix~\ref{sec:appendix-prompts} and the use of panel verdicts as reliable measurements of legal analysis and professional intake quality.

\end{document}